\documentclass[lettersize,journal]{IEEEtran}
\usepackage{amsmath,amsfonts}
\usepackage{algorithmic}
\usepackage{algorithm}
\usepackage{array}
\usepackage[caption=false,font=normalsize,labelfont=sf,textfont=sf]{subfig}
\usepackage{textcomp}
\usepackage{stfloats}
\usepackage{url}
\usepackage{verbatim}
\usepackage{graphicx}
\usepackage{cite}
\usepackage{threeparttable}
\usepackage{tablefootnote}

\usepackage{amsthm}
\usepackage[dvipsnames]{xcolor}
\usepackage{siunitx}
\usepackage{epstopdf}
\usepackage{svg}
\usepackage{multirow}
\usepackage{booktabs}
\usepackage{comment}
\usepackage{cuted}
\usepackage{lipsum}
\usepackage{txfonts}
\usepackage{balance}

\usepackage{nccmath}

\theoremstyle{definition}
\newtheorem{theorem}{Theorem}

\newtheorem{proposition}{Proposition}
\newtheorem{corollary}{Corollary}
\newtheorem{definition}{Definition}
\newtheorem{assumption}{Assumption}
\newtheorem{remark}{Remark}

\begin{document}

\title{Multi-Robot Coordination via Stochastic Dynamic Games with\\Strategic Decentralized Sparse Estimation}

\author{
Tianyu Qiu, Filippos Fotiadis, Ufuk Topcu and David Fridovich-Keil
\thanks{T. Qiu, F. Fotiadis, U. Topcu, and D. Fridovich-Keil are with the Oden Institute for Computational Engineering and Sciences, The University of Texas at Austin, TX, 78712, USA. 
(Email:  {\tt \{tianyuqiu, ffotiadis, utopcu, dfk\}@utexas.edu}).
}%
}




\maketitle

\begin{abstract}
Multi-robot systems often coordinate using individual, noisy state observations under sensing limitations. 
Stochastic dynamic games model these robot interactions, but the coupling between control decisions and individual state estimates makes computing equilibrium strategies challenging. 
We study this problem in $N$-robot linear-quadratic Gaussian (LQG) games with noisy individual observations.
We derive first-order necessary conditions for local Nash equilibrium within the affine feedback class of strategies for fixed linear estimators.
Then, we use these conditions in an alternating control–estimation iteration to compute affine feedback equilibrium policies and stepwise minimum-variance estimators.
We extend the framework to smooth nonlinear games through iterative linear-quadratic (iLQ) approximations. 
To reduce sensor usage per robot, we leverage a policy-adaptive group-lasso regularization scheme, assigning smaller group-lasso penalties to estimators associated with larger feedback-gain blocks.
It allows robots to allocate observations according to their control needs. 
Across multi-robot tasks, our method reduces mean robot cost by up to $40.0\%$ and task-specific metrics (tracking error in formation and minimum separation in navigation) by up to $52.7\%$ vs. deterministic Nash baselines.
Policy-adaptive sparse estimation reduces the need for sensor usage by up to $79.2\%$, at the expense of a $4.2\%$ increase in robot cost compared to dense estimation, offering a tunable trade-off between robot cost and sensor usage.
\end{abstract}

\begin{IEEEkeywords}
Optimization and Optimal Control, Multi-Robot Sytems, Motion and Path Planning, Stochastic Dynamic Games
\end{IEEEkeywords}

\section{Introduction}
\label{sec:introduction}
Multi-robot systems must often make decisions under process and sensing uncertainty. 
When each robot in such a system has a different objective, the resulting decision problem can be formulated as a stochastic dynamic game whose Nash equilibria are challenging to compute. 
Moreover, resource constraints often limit robots' ability to communicate and exchange measurements with one another. 
For example, the NASA Cooperative Autonomous Distributed Robotic Exploration (CADRE) mission \cite{saboia2024cadre} consists of a group of small rovers which must coordinate their movements autonomously to map the lunar surface, while satisfying a tight energy budget that limits communication.

\begin{figure}[t]
    \centering
    \includegraphics[width=1\linewidth]{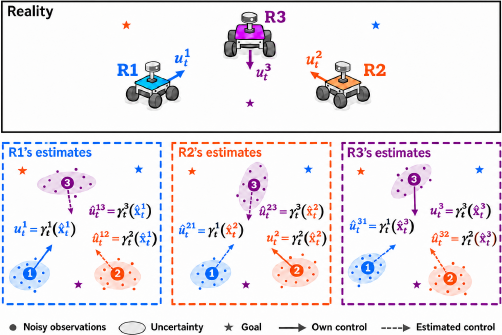}
    \caption{Multi-robot planning under individual observations with no communication.
    Robots share dynamics models, precomputed policies, and estimators before deployment but do not exchange measurements, estimates, or realized controls during execution.
    Each robot $i\in\{1,2,3\}$ receives an individual sensor observation, forms a private estimate of the joint state of the multi-robot system $\hat{\textbf{x}}_t^i$, and computes control $u_t^i$. 
    Since state estimates are private, each robot can only predict others' control inputs using its own estimate $\hat{\textbf{x}}_t^i$. 
    This results in a mismatch between predicted and realized joint controls, making the planning problem challenging.}
    \label{fig:front}
\end{figure}

Motivated by communication-constrained missions such as CADRE, we study stochastic dynamic games in which robots cannot exchange measurements, estimates, or realized controls online. Instead, each robot must rely on its own noisy observations to estimate the joint state and determine its control actions. We refer to this knowledge pattern as decentralized estimation and decision-making, as illustrated in Fig.~\ref{fig:front}.

This setting presents two challenges. 
First, decentralized estimation couples control and estimation: each robot acts on its own state estimate, while the resulting joint actions affect the physical state and future estimation errors of every robot. 
Feedback design must therefore account for the uncertainty induced by individual observations. 
Second, limited sensing resources prevent each robot from using all available observation groups. 
At each stage, each robot must select measurement blocks that support its control decisions, as omitting observations changes its estimation errors and can affect the costs incurred by all robots.

We address these challenges in $N$-robot general-sum linear-quadratic Gaussian (LQG) games with decentralized estimation. 
Building on the sparse-estimation framework of Qiu et al.~\cite{qiu2026linear}, which uses fixed deterministic feedback policies, we incorporate estimation-error moments into policy computation and apply policy-dependent regularization for subsequent sparse estimation.

Our contributions are:

\begin{enumerate}
    \item \textbf{Analytical robot cost characterization and affine feedback Nash conditions.}
    We derive a quadratic expression for each robot's expected cost under affine feedback policies and linear estimators.
    Based on the cost characterization, we derive first-order necessary conditions and a sufficient condition for local feedback Nash equilibrium within the affine policy class under a set of fixed linear state estimators.
    \item \textbf{Iterative control–estimation optimization.}
    We develop an alternating optimization framework for affine Nash policy computation and stepwise minimum-variance estimation, and evaluate satisfaction of the policy stationarity equations using numerical residuals.
    \item \textbf{Policy-adaptive sparse estimation.}
    To reduce sensor use while preserving important measurements for decision-making, we adopt the group-lasso estimator structure of Qiu et al.~\cite{qiu2026linear}.
    Sparse measurement selection is guided by the computed feedback policies and is evaluated with those policies fixed.
    We evaluate the trade-off between robot cost and sensor use with this cost characterization introduced above.
    \item \textbf{Extension to nonlinear games and robotic systems.}
    We extend the computation to smooth nonlinear games and evaluate it on CADRE-inspired planar formation and navigation, drone formation, and manipulation tasks, together with a two-Jackal navigation demonstration. 
    Across the tested configurations, our method reduces mean realized cost by up to $40.0\%$ and improves task-specific metrics by up to $52.7\%$ relative to deterministic Nash baselines.
    With the dense feedback policy held fixed, policy-adaptive sparse estimation achieves a trade-off of up to $79.2\%$ fewer measurement blocks for a $4.2\%$ increase in analytical cost relative to dense estimation.
\end{enumerate}

\section{Related Work}
\label{sec:related_work}
Existing work on planning under noise mostly considers 1) single-agent problems (Sec.~\ref{sec:LQG}), which do not involve strategic interactions among robots, or 2) stochastic games with shared observations and a common state estimate (Sec.~\ref{sec:sg_ci}), which do not account for the mismatch among agents' private estimates considered here.
In terms of sparse sensing, existing methods typically select measurements to improve estimation accuracy or a shared control objective; the closely related decentralized game framework of Qiu et al.~\cite{qiu2026linear} instead uses fixed feedback policies computed from a deterministic game.
Our work accounts for private estimation uncertainty in feedback policy computation and subsequently uses the resulting policies to guide sparse measurement selection.
The following subsections compare these information structures and sensing objectives.

\subsection{Single-Agent LQG}
\label{sec:LQG}
Classical linear-quadratic Gaussian (LQG) control for single-agent systems \cite{bertsekas2012dynamic,lewis2017optimal} admits a separation principle that permits independent controller and estimator design: the optimal feedback gain is the deterministic linear-quadratic regulator (LQR) gain, and the optimal linear estimator is the Kalman filter. However, this separation principle need not hold in decentralized information settings
 \cite{witsenhausen1968counterexample}.
In our game, decentralized information takes the form of individual observations. Specifically, each agent updates a private estimate of the joint system's state, and its estimate-dependent action affects the state and the other agents' estimation errors.

\subsection{Stochastic Games with Shared Belief}
\label{sec:sg_ci}
A common approach to stochastic dynamic games under process and sensing uncertainty is to formulate strategies using common information or shared belief states \cite{gupta2014common,gupta2016dynamic,schwarting2021stochastic}. Specifically, when agents have access to common observations, uncertainty can be represented by a shared belief propagated through a common filtering recursion, such as a Kalman or extended Kalman filter. However, maintaining such shared beliefs generally relies on common observations or publicly available signals, which may be limited in communication-constrained robotic systems.

Related linear-quadratic (LQ) game formulations assume control sharing or other publicly observed signals that support common belief updates ~\cite{pachter2016linear,pachter2017lqg}.
Under decentralized estimation, by contrast, each agent maintains an individual state estimate from private observations.

\subsection{Stochastic Games with Partial Information}
\label{sec:sg_pi}
Prior work studies general-sum asymmetric-information games through structured perfect Bayesian equilibria, using public information or observed actions to correct beliefs about other agents ~\cite{vasal2016signaling,heydaribeni2019linear,hambly2023linear}.
Forward-backward and best-response methods have also been developed for LQG games with partial and asymmetric observations \cite{guan2026forward,guan2025best}, but the cited formulations focus on two-agent zero-sum games.

Our formulation allows $N$ agents with general-sum objectives, where each agent maintains a private estimate of the joint state without exchanging observations or realized controls online.
For fixed linear estimators, we derive stationarity conditions for affine feedback policies and establish a sufficient condition for local Nash equilibrium within this policy class.

\subsection{Sparse Sensing for Resource-Aware Multi-Robot Coordination}
\label{sec:sparse_sensing}
Sparse sensor selection and control-sensing co-design seek to balance sensing resources against estimation accuracy or control performance \cite{tzoumas2020lqg,zhong2024sparse,fotiadis2024input,fotiadis2026input}. In multi-robot systems, related approaches address sensing and communication constraints through selective routing of observations \cite{kassir2015communication}, periodic coordination of sensing and motion \cite{lauri2017periodic}, and decentralized active perception with intermittent exchange of planning information \cite{best2019decmcts}. These approaches manage limited sensing and communication resources to support estimation and coordination, but do not directly address the coupling between feedback policies and private state estimation in stochastic dynamic games.

More closely related to our work, Qiu et al.~\cite{qiu2026linear} combine decentralized group-lasso estimation with fixed feedback Nash policies computed from a deterministic LQ game \cite{bacsar1998dynamic}. 
In contrast, we account for estimation uncertainty when computing feedback policies, alternating backward policy updates with forward minimum-variance estimator updates. 
Following \cite{qiu2026linear}, we hold the resulting policies fixed and apply policy-adaptive group-lasso regularization that favors measurement groups associated with stronger feedback dependence. We evaluate the resulting trade-off between sensing resource use and expected game cost.

\subsection{Iterative LQ (iLQ) Methods}
Differential dynamic programming~\cite{tassa2014control} and iterative LQR~\cite{li2004iterative} solve nonlinear optimal-control problems through local approximations; iterative LQ game solvers extend this idea to multi-agent objectives~\cite{fridovich2020efficient,laine2023computation}, while belief-space methods account for state uncertainty~\cite{schwarting2021stochastic}.
We use this local approximation technique to extend the affine control–estimation computation to smooth nonlinear stochastic games with individual observations.

\section{Stochastic Dynamic Games with Individual Observations}
\label{sec:problem}
\subsection{Stochastic Dynamic Games}
We consider an $N$-robot discrete-time stochastic dynamic game over a finite horizon of length $T\in\mathbb{N}$.
Let $[N]\coloneqq\{1,\dots,N\}$. Each robot $i\in[N]$ has state $x_t^i\in\mathbb{R}^{n_i}$ and control input $u_t^i\in\mathbb{R}^{m_i}$.
The state evolves subject to stochastic dynamics:

\begin{equation*}
        x_{t+1}^i\coloneqq f_t^i(x_t^i,u_t^i)+w_t^i,\qquad t\in[T],
\end{equation*}
where $f_t^i\colon\mathbb{R}^{n_i}\times\mathbb{R}^{m_i}\to\mathbb{R}^{n_i}$ is a smooth map and the process disturbance $w_t^i\in\mathbb{R}^{n_i}$ is Gaussian, $w_t^i\sim\mathcal{N}(0,W_t^i)$, with covariance $W_t^i\in\mathbb{R}^{n_i\times n_i}\succeq0$.

We write the joint state and joint control for all robots as ${\textbf{x}_t\coloneqq[x_t^{1\top},\dots,x_t^{N\top}]^\top\in\mathbb{R}^n}$ and ${\textbf{u}_t\coloneqq[u_t^{1\top},\dots,u_t^{N\top}]^\top\in\mathbb{R}^m}$, respectively, where ${n\coloneqq\sum_{i=1}^N n_i}$ and ${m\coloneqq\sum_{i=1}^N m_i}$.
Stacking the individual dynamics yields the joint evolution model

\begin{equation*}
    \textbf{x}_{t+1}=f_t(\textbf{x}_t,\textbf{u}_t)+\textbf{w}_t,
\end{equation*}
where the joint map $f_t\colon\mathbb{R}^n\times\mathbb{R}^m\to\mathbb{R}^n$ is defined by

\begin{equation*}
    f_t(\textbf{x}_t,\textbf{u}_t)
    \coloneqq
    \left[f_t^{1}(x_t^1,u_t^1)^\top,\dots,f_t^{N}(x_t^N,u_t^N)^\top\right]^\top,
\end{equation*}
and the joint process noise $\textbf{w}_t\coloneqq[w_t^{1\top},\dots,w_t^{N\top}]^\top\in\mathbb{R}^n$ satisfies

\begin{equation*}
    \textbf{w}_t\sim\mathcal{N}(\textbf{0},W_t),\qquad
    W_t\coloneqq\mathrm{blockdiag}(\{W_t^i\}_{i=1}^N)\in\mathbb{R}^{n\times n}.
\end{equation*}
Each robot~$i$ seeks to minimize an expected cumulative cost $\mathcal{J}^i$.
Letting $\textbf{u}^i\coloneqq\{u_t^i\}_{t=1}^{T}$ denote robot~$i$'s control sequence, the problem of robot~$i$ is

\begin{subequations}
    \begin{align}
        \min_{\textbf{u}^i}\ &\underbrace{\mathbb{E}\left[\sum_{t=1}^{T}\ell_t^i(\textbf{x}_t,\textbf{u}_t)+\ell_{T+1}^i(\textbf{x}_{T+1})\right]}_{\mathcal{J}^i},\label{eqn:dgg_cost}\\
        \text{s.t.}\  \textbf{x}_{t+1}&=f_t(\textbf{x}_t,\textbf{u}_t)+\textbf{w}_t,\label{eqn:dgg_dyn}\\
        \textbf{x}_1&\sim\mathcal{N}(\Bar{\textbf{x}}_1,\Sigma_{\textbf{x},1}^-),\label{eqn:dgg_initial_cond}
    \end{align}
    \label{eqn:DTDGG}%
\end{subequations}
where $\ell_t^i\colon\mathbb{R}^n\times\mathbb{R}^m\to\mathbb{R}$ are smooth stage costs, $\ell_{T+1}^i\colon\mathbb{R}^n\to\mathbb{R}$ is a smooth terminal cost, and the joint initial state $\textbf{x}_1$ follows a Gaussian prior distribution \eqref{eqn:dgg_initial_cond} with mean $\Bar{\textbf{x}}_1$ and covariance $\Sigma_{\textbf{x},1}^-\succeq0$.
The stage costs $\ell_t^i$ depend on the joint state and joint control, and the terminal costs $\ell_{T+1}^i$ depend on the joint state.
The $N$ instances of~\eqref{eqn:DTDGG} are coupled and together constitute a stochastic dynamic game.

\begin{figure*}[t]
    \centering
    \includegraphics[width=1\linewidth]{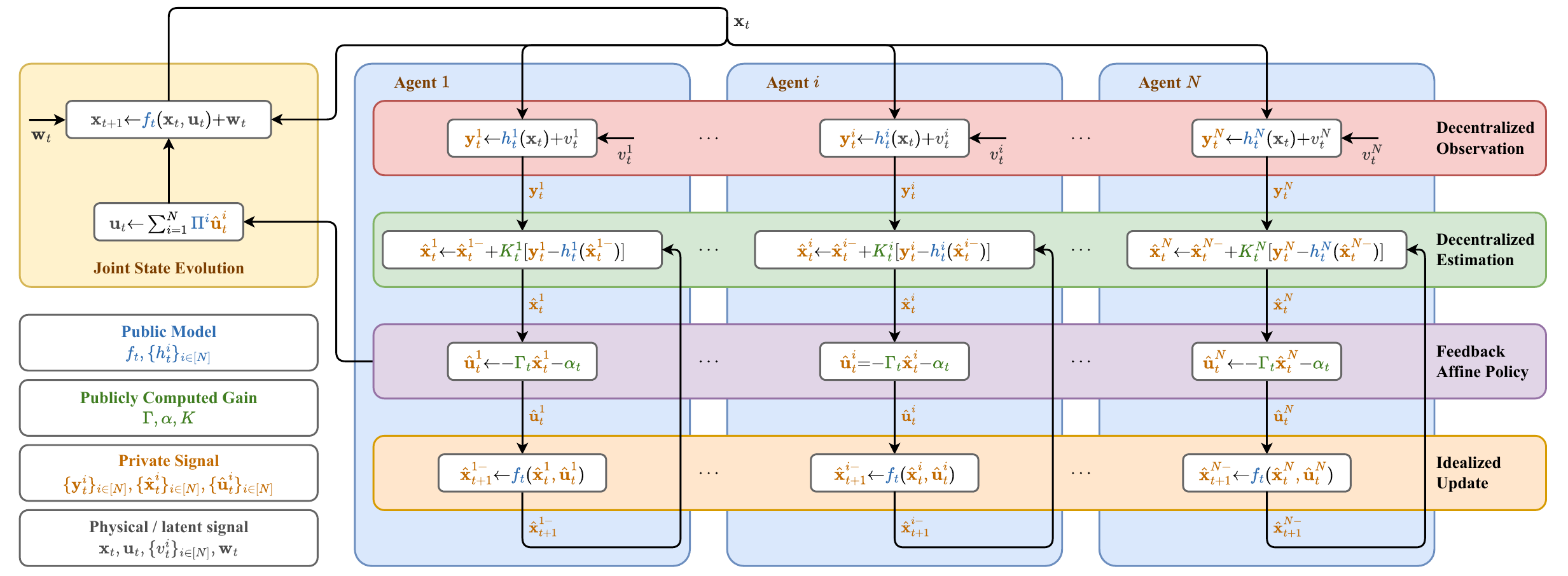}
    \caption{Knowledge pattern and information flow under decentralized estimation. Each robot $i$ receives an individual observation $\textbf{y}_t^i$, forms an individual estimate $\hat{\textbf{x}}_t^i$, and applies an affine feedback policy, producing an imputed joint input $\hat{\textbf{u}}_t^i$. The model and the offline-computed policies $(\Gamma_t,\alpha_t)$ and estimators $\{K_t^i\}$ are public; $\textbf{y}_t^i$, $\hat{\textbf{x}}_t^i$, and $\hat{\textbf{u}}_t^i$ remain private. Robot $i$ forms the next prior state estimate $\hat{\textbf{x}}_{t+1}^{i-}$ with $\hat{\textbf{u}}_t^i$, while the stacked realized controls jointly advance the state to $\textbf{x}_{t+1}$, which couples the feedback policies and estimators through the closed-loop state and estimation-error dynamics.}
    \label{fig:decentralized_information_flow}
\end{figure*}

\subsection{Decentralized Estimation with Individual Observations}
In practice, each robot~$i$ does not, in general, observe the full joint state~$\textbf{x}_t$.
Instead, it receives a partial, noisy individual observation $\textbf{y}_t^i\in\mathbb{R}^{p_i}$ generated by

\begin{equation}
        \textbf{y}_t^i\coloneqq h_t^i(\textbf{x}_t)+v_t^i,\label{eqn:ind_obs}
\end{equation}
where $h_t^i\colon\mathbb{R}^n\to\mathbb{R}^{p_i}$ is a smooth observation map and $v_t^i\in\mathbb{R}^{p_i}$ is Gaussian observation noise, $v_t^i\sim\mathcal{N}(0,V_t^i)$, with covariance $V_t^i\in\mathbb{R}^{p_i\times p_i}\succeq0$.
Concatenating all robots' individual  noises yields $\textbf{v}_t\coloneqq[v_t^{1\top},\dots,v_t^{N\top}]^\top\in\mathbb{R}^p$, where $p\coloneqq\sum_{i=1}^N p_i$ and

\begin{equation*}
    \textbf{v}_t\sim\mathcal{N}(\textbf{0},V_t),\qquad
    V_t\coloneqq\mathrm{blockdiag}(\{V_t^i\}_{i=1}^N)\in\mathbb{R}^{p\times p}.
\end{equation*}
We assume that the process and observation noises are mutually independent across robots and time and are independent of the initial state.

\subsection{Individual Estimation and Feedback Control}
Each robot updates its estimate of the joint state using a linear estimator $K_t^i\in\mathbb{R}^{n\times p_i}$:

\begin{equation*}
    \hat{\textbf{x}}_t^i\coloneqq \hat{\textbf{x}}_t^{i-}+K_t^i\left[\textbf{y}_t^i-h_t^i(\hat{\textbf{x}}_t^{i-})\right],
\end{equation*}
which corrects the state prediction $\hat{\textbf{x}}_t^{i-}$ using its individual observation $\textbf{y}_t^i$. Robot $i$ then computes its feedback control input as

\begin{equation*}
    u_t^i\coloneqq\gamma_t^i(\hat{\textbf{x}}_t^i),
\end{equation*}
with feedback strategy $\gamma_t^i$.

\subsection{Knowledge Pattern}
In this work, we seek a feedback Nash strategy for the stochastic dynamic game~\eqref{eqn:DTDGG} under individual observations~\eqref{eqn:ind_obs}. 
The following assumption specifies which quantities are shared and which remain private.

\begin{assumption}[Knowledge pattern]
\label{ass:info_pattern}
At each stage $t\in[T]$, robots have the following common and private knowledge:

\begin{enumerate}
    \item \textbf{Common knowledge.} The robots’ costs $\mathcal{J}^i$, the system and observation maps $f_t, \{h_t^i\}_{i\in[N]}$, the noise covariances $W_t, V_t$, and the initial prior $(\Bar{\textbf{x}}_1,\Sigma_{\textbf{x},1}^-)$ are common knowledge among all robots. 
    \item \textbf{Private knowledge.} Each robot~$i$'s individual observation $\textbf{y}_t^i$, together with the estimate $\hat{\textbf{x}}_t^i$ and the control $u_t^i$ computed from it, is private and is \textbf{not} communicated among robots.
    \item \textbf{Policies and estimators.} The feedback strategies $\{\gamma_t^i\}_{i\in[N]}$ and linear estimators $K$ are optimized from this common knowledge and are shared among all robots. 
\end{enumerate}
\end{assumption}
To predict its next state estimate $\hat{\textbf{x}}_{t+1}^{i-}$ using \eqref{eqn:dgg_dyn}, robot $i$ needs the joint control $\textbf{u}_t$, including the other robots' actual control inputs.
However, $\textbf{u}_t$ is not available since robots do not communicate. 
The realized joint control stacks inputs computed from the robots' different private estimates.
Therefore, we adopt the following prediction approximation: robot $i$ substitutes its own estimate $\hat{\textbf{x}}_t^i$ for the other robots' estimates $\hat{\textbf{x}}_t^j (j\neq i)$.
We summarize this as the following first-order self-projecting belief assumption, which has also been used in \cite{guan2026forward,schwarting2021stochastic,qiu2026linear}:

\begin{assumption}[First-order self-projecting beliefs]
\label{ass:first_order_belief}
Robot~$i$ assumes that every other robot $j\neq i$ forms the same estimate of the joint state as robot~$i$, i.e., robot~$i$ imputes $\hat{\textbf{x}}_t^j\leftarrow\hat{\textbf{x}}_t^i$. 
\end{assumption}
Robot $i$ then estimates the other robots' control inputs based on this assumption, i.e.,

\begin{equation*}
    \hat{u}_t^{ij}\coloneqq\gamma_t^j(\hat{\textbf{x}}_t^i),\quad
    \hat{\textbf{u}}_t^i\coloneqq\gamma_t(\hat{\textbf{x}}_t^i),
\end{equation*}
where $\gamma_t(\cdot)\coloneqq[\gamma_t^1(\cdot)^\top,\ldots,\gamma_t^N(\cdot)^\top]^\top$ is the map from joint state to feedback control inputs.
Fig.~\ref{fig:decentralized_information_flow} illustrates this process.
The joint control predicted from robot $i$'s estimate generally differs from the realized joint control, as shown in Fig.~\ref{fig:front}. 
This information structure creates a mismatch between each robot’s estimate of the joint input $\hat{\textbf{u}}_t^i$ and the realized joint input $\textbf{u}_t$, and thus makes the search for an equilibrium substantially more challenging.

\subsection{Feedback Nash Equilibrium}
With the knowledge pattern of Assumption~\ref{ass:info_pattern}, we seek a local feedback Nash equilibrium of~\eqref{eqn:DTDGG}.
Collect robot~$i$'s stagewise strategies into $\gamma^i\coloneqq\{\gamma_t^i\}_{t=1}^{T}$, the joint profile into $\gamma\coloneqq\{\gamma^i\}_{i\in[N]}$, and the other robots' stage-$t$ strategies into $\gamma_t^{-i}\coloneqq\{\gamma_t^j\}_{j\neq i}$.
The expected cost $\mathcal{J}^i(\gamma)$ is determined by the closed-loop trajectories induced via~\eqref{eqn:dgg_dyn}.

\begin{definition}[Local feedback Nash equilibria]
\label{def:local_nash}
Let $\mathcal G_t^i$ be robot $i$'s admissible feedback-policy space at stage $t$, equipped with a metric $d_t^i$. A strategy profile $\gamma^\star$ is a \emph{local feedback Nash equilibrium} if, for every robot $i\in[N]$, stage $t\in[T]$, and past strategy profile $\gamma_{1:t-1}$, there exists $\varepsilon>0$ such that

\begin{equation*}
    \mathcal J^i(\gamma_{1:t-1},\gamma_{t:T}^\star)
    \leq
    \mathcal J^i(\gamma_{1:t-1},\gamma_t^i,\gamma_t^{-i\star},\gamma_{t+1:T}^\star)
\end{equation*}
for every $\gamma_t^i\in\mathcal G_t^i$ satisfying $d_t^i(\gamma_t^i,\gamma_t^{i\star})<\varepsilon$. Here $\gamma_{a:b}$ denotes the collection of all robots' strategies over stages $a$ through $b$. Thus, no robot can reduce its expected cost through a sufficiently small unilateral change in its stage-$t$ feedback policy, with the other robots' policies and all continuation policies fixed.
\end{definition}
Feedback Nash equilibria are well studied under perfect-state feedback~\cite{SmoothGameTheory} and under shared observations~\cite{schwarting2021stochastic}.
Under Assumption~\ref{ass:info_pattern}, finding closed-form solutions remains an open problem; the next section develops a computational procedure for this setting.

\begin{remark}[Affine policy restriction]
\label{rmk:affine}
    Though we define the feedback Nash equilibrium for arbitrary feedback maps from estimates to controls, we emphasize that it is fundamentally intractable to find a closed-loop solution without any presumption about the policy class to the best of our knowledge.
    Therefore, throughout the remainder of the paper, we restrict each robot to an affine policy to solve the stochastic game with individual observations. This is motivated by the fact that the optimal policy for a single-agent LQG problem and an equilibrium policy for a deterministic LQ game belong to the affine class (though for these two problems we do not need to impose this policy restriction).
\end{remark}

\subsection{Problem Statement}
\label{sec:problem_statement}
Following Remark \ref{rmk:affine}, we seek local feedback Nash policies in the affine policy class for stochastic dynamic games with individual observations.
Each robot applies a policy that is affine in its joint-state estimate $\hat{\textbf{x}}_t^i$:

\begin{equation}
u_t^i=\gamma_t^i(\hat{\textbf{x}}_t^i)=-\Gamma_t^i\hat{\textbf{x}}_t^i-\alpha_t^i.\label{eqn:actual_single_ctrl}
\end{equation}
Here $\Gamma_t^i\in\mathbb R^{m_i\times n}$ is robot $i$'s feedback gain and $\alpha_t^i\in\mathbb R^{m_i}$ is its affine offset.
We stack the gains $\Gamma_t$ and affine terms $\alpha_t$ as

\begin{equation*}
\Gamma_t\coloneqq\begin{bmatrix}\Gamma_t^{1\top}&\cdots&\Gamma_t^{N\top}\end{bmatrix}^\top,\qquad
\alpha_t\coloneqq\begin{bmatrix}\alpha_t^{1\top}&\cdots&\alpha_t^{N\top}\end{bmatrix}^\top,
\end{equation*}
so that $\Gamma_t^i$ is the $i$-th block row of $\Gamma_t$, and $\alpha_t^i$ is the $i$-th block of $\alpha_t$.
We write $(\Gamma,\alpha)\coloneqq(\Gamma_{1:T},\alpha_{1:T})$ for these sequences over the horizon, and $K\coloneqq\{K_{1:T+1}^i\}_{i\in[N]}$ for the estimation-gain sequences.
For a single stage $t$, we write $(\Gamma_t,\alpha_t)$ and $\{K_t^i\}$.
Our objective now is to find local affine feedback Nash strategies $(\Gamma,\alpha)$ and estimators $K$ that minimize the trace of each robot's stepwise estimation covariance.
The gains $K$ determine the recursive estimates $\hat{\textbf{x}}_t^i$, and $(\Gamma,\alpha)$ map these estimates to the realized controls through $u_t^i=-\Gamma_t^i\hat{\textbf{x}}_t^i-\alpha_t^i$. Thus, we optimize a parameterized class of observation-dependent control strategies for~\eqref{eqn:DTDGG}.

In Sec.~\ref{sec:decentralized_est_gain}, we further sparsify $K$ to permit selected measurement groups to be dropped while holding the feedback policies fixed, and evaluate the resulting change in expected game cost. The policy and estimation gains are computed offline and shared before deployment, while observations, estimates, and realized controls remain private online.

\section{Computation of Affine Feedback Nash Strategies}
\label{sec:method}
\subsection{Linear-Quadratic Setup}
Following relevant works \cite{guan2026forward,qiu2026linear}, we begin with the affine-quadratic specialization of \eqref{eqn:DTDGG}-\eqref{eqn:ind_obs}, and subsequently extend to general nonlinear stochastic games in Sec.~\ref{sec:joint_ilq_alt}.
In this setting, robot~$i$'s stage and terminal costs, the joint dynamics, and the individual observation models are

\begin{equation*}
    \begin{aligned}
        \ell_t^i(\textbf{x}_t,\textbf{u}_t)
        &=\ \frac{1}{2}\textbf{x}_t^\top Q_t^i\textbf{x}_t+q_t^{i\top}\textbf{x}_t
        +\frac{1}{2}\textbf{u}_t^\top R_t^i\textbf{u}_t+r_t^{i\top}\textbf{u}_t
        +\ell_{\mathrm{const},t}^i,\\
        \ell_{T+1}^i(\textbf{x}_{T+1})
        &=\ \frac{1}{2}\textbf{x}_{T+1}^\top Q_{T+1}^i\textbf{x}_{T+1}
        +q_{T+1}^{i\top}\textbf{x}_{T+1}+\ell_{\mathrm{const},T+1}^i,\\
        \textbf{x}_{t+1}
        &=\ A_t\textbf{x}_t+B_t\textbf{u}_t+f_{\mathrm{const},t}+\textbf{w}_t,\\
        \textbf{y}_t^i
        &=\ C_t^i\textbf{x}_t+h_{\mathrm{const},t}^i+v_t^i,
    \end{aligned}
\end{equation*}
where $Q_t^i\succeq0$, $R_t^i\succ0$, and the residuals $\ell_{\mathrm{const},t}^i$, $f_{\mathrm{const},t}$, and $h_{\mathrm{const},t}^i$ are known stagewise constants.
Partition the joint input matrix by robot as

\begin{equation*}
    B_t\coloneqq\begin{bmatrix}B_t^1&\cdots&B_t^N\end{bmatrix}
    \in\mathbb{R}^{n\times m},\qquad B_t^i\in\mathbb{R}^{n\times m_i}.
\end{equation*}
The block $B_t^i$ maps robot $i$'s input into the joint-state dynamics.

\subsection{Affine Strategies and System Evolution}
\label{sec:affine_class}
Using the affine policies and gain notation defined in Sec.~\ref{sec:problem_statement}, we now derive the joint state and estimation-error dynamics.
Write $\textbf{e}_t^i\coloneqq\textbf{x}_t-\hat{\textbf{x}}_t^i$, $\textbf{e}_t\coloneqq[\textbf{e}_t^{1\top},\dots,\textbf{e}_t^{N\top}]^\top$, $\Sigma_t^i\coloneqq\operatorname{Cov}(\textbf{e}_t^i)$, and $\Sigma_t\coloneqq\operatorname{Cov}(\textbf{e}_t)$.
Under Assumption~\ref{ass:first_order_belief}, robot~$i$ \textit{imputes} the joint input as

\begin{equation}
    \hat{\textbf{u}}_t^i=-\Gamma_t\hat{\textbf{x}}_t^i-\alpha_t=-\Gamma_t(\textbf{x}_t-\textbf{e}_t^i)-\alpha_t,
    \label{eqn:estimated_control}%
\end{equation}
and $u_t^i$ is the $i$-th block of $\hat{\textbf{u}}_t^i$. 
The actual joint control is the concatenation of privately computed inputs, i.e.,

\begin{equation}
    \textbf{u}_t=\sum_{i=1}^N \Pi^i\hat{\textbf{u}}_t^i
    =-\alpha_t-\Gamma_t\textbf{x}_t+\sum_{i=1}^N \Pi^i\Gamma_t\textbf{e}_t^i,
    \label{eqn:actual_control}
\end{equation}
where $\Pi^i\in\mathbb{R}^{m\times m}$ embeds robot~$i$'s input into the stacked $\textbf{u}_t$: it is block diagonal with $I_{m_i}$ in the $(i,i)$ block and zeros elsewhere.
The joint state therefore evolves as

\begin{equation}
    \textbf{x}_{t+1}
    =(A_t-B_t\Gamma_t)\textbf{x}_t+\sum_{i=1}^N B_t \Pi^i\Gamma_t\textbf{e}_t^i-B_t\alpha_t+f_{\mathrm{const},t}+\textbf{w}_t.
    \label{eqn:x_update}
\end{equation}
Each robot predicts the next state using the imputed joint input $\hat{\textbf{u}}_t^i$ and corrects that prediction using its private innovation:

\begin{subequations}
    \begin{align}
        \hat{\textbf{x}}_{t+1}^{i-}&=A_t\hat{\textbf{x}}_t^i+B_t\hat{\textbf{u}}_t^i+f_{\mathrm{const},t},\label{eqn:prior_x_update}\\
        \hat{\textbf{x}}_{t+1}^i&=\hat{\textbf{x}}_{t+1}^{i-}+K_{t+1}^i\left[\textbf{y}_{t+1}^i-C_{t+1}^i\hat{\textbf{x}}_{t+1}^{i-}-h_{\mathrm{const},t+1}^i\right]\notag\\
        &=\hat{\textbf{x}}_{t+1}^{i-}+K_{t+1}^i\left[C_{t+1}^i(\textbf{x}_{t+1}-\hat{\textbf{x}}_{t+1}^{i-})+v_{t+1}^i\right].\label{eqn:post_x_update}
    \end{align}
    \label{eqn:est_x_update}%
\end{subequations}
Below, Sec.~\ref{sec:decentralized_est_gain} develops both dense minimum-variance and sparse alternatives for the linear estimation gain $K_t^i$.
Subtracting~\eqref{eqn:est_x_update} from~\eqref{eqn:x_update} yields the following error recursion for each robot $i$, in which 
the terms multiplied by $B_t$ capture the mismatch between realized and imputed joint controls:

\begin{subequations}
    \begin{align}
        \textbf{e}_{t+1}^{i-}
        &=A_t\textbf{e}_t^i+B_t\left(\sum_{j=1}^N \Pi^j\Gamma_t\textbf{e}_t^j-\Gamma_t\textbf{e}_t^i\right)+\textbf{w}_t,\label{eqn:ind_err_prior_update}\\
        \textbf{e}_{t+1}^i
        &=\textbf{e}_{t+1}^{i-}-K_{t+1}^i\left(C_{t+1}^i\textbf{e}_{t+1}^{i-}+v_{t+1}^i\right).\label{eqn:ind_err_post_update}
    \end{align}
    \label{eqn:est_update}%
\end{subequations}
The mismatch due to imputation couples the robots' estimation errors, so their cross-covariances must be propagated jointly.

\subsection{Affine Nash Strategy via Dynamic Programming}
To solve for the affine Nash strategy, we first introduce the augmented state $\textbf{z}_t\coloneqq[\textbf{x}_t^\top,\textbf{e}_t^\top]^\top$ and the augmented noise $\omega_t\coloneqq[\textbf{w}_t^\top,\textbf{v}_{t+1}^\top]^\top\sim\mathcal N(\textbf{0},\Omega_t)$, where $\Omega_t=\text{diag}\{W_t,V_{t+1}\}$.
Index the $N+1$ blocks of $\textbf{z}_t$ by $0,1,\ldots,N$, with block $0$ equal to $\textbf{x}_t$ and block $i$ equal to $\textbf{e}_t^i$. Each $E_i\in\mathbb R^{n\times(N+1)n}$ has $I_n$ in block $i$ and $\textbf{0}_{n\times n}$ elsewhere.

\begin{equation*}
    \begin{aligned}
        E_0&\coloneqq E_{\textbf{x}}\coloneqq\begin{bmatrix}I_n&0&\cdots&0\end{bmatrix},\\
    E_i&\coloneqq\begin{bmatrix}0&\cdots&0&I_n&0&\cdots&0\end{bmatrix},\quad i\in[N].
    \end{aligned}
\end{equation*}
The matrices $E_i$ act as selectors with $\textbf{x}_t=E_0\textbf{z}_t$ and $\textbf{e}_t^i=E_i\textbf{z}_t$.
Set

\begin{equation}
D_{ab}\coloneqq E_a-E_b
\label{eqn:Dis}
\end{equation}
for indices $a,b\in\{0\}\cup[N]\ (a\neq b)$.
It follows that

\begin{equation*}
\hat{\textbf{x}}_t^i=\textbf{x}_t-\textbf{e}_t^i=D_{0i}\textbf{z}_t,\qquad
\textbf{e}_t^j-\textbf{e}_t^\ell=D_{j\ell}\textbf{z}_t.
\end{equation*}
The affine strategies \eqref{eqn:actual_single_ctrl} then become

\begin{equation}
    \begin{aligned}
        u_t^i&=-\Gamma_t^i\hat{\textbf{x}}_t^i-\alpha_t^i=-\Gamma_t^i D_{0i}\textbf{z}_t-\alpha_t^i,\\
        \textbf{u}_t&=-\alpha_t-\sum_{j=1}^N \Pi^j\Gamma_t D_{0j}\textbf{z}_t=-\Gamma_{\textbf{z},t}\textbf{z}_t-\alpha_t,
    \end{aligned}
    \label{eqn:u_from_z}
\end{equation}
Here $\Gamma_{\textbf{z},t}\coloneqq\sum_{j=1}^N \Pi^j\Gamma_t D_{0j}$ is the joint feedback gain matrix for the augmented state $\textbf{z}_t$.
Substituting \eqref{eqn:u_from_z} into the linear dynamics and the prior/posterior error recursions \eqref{eqn:x_update}-\eqref{eqn:est_update} and stacking the rows yields the closed-loop augmented dynamics

\begin{equation}
    \begin{aligned}
        \textbf{z}_{t+1}
        &=\underbrace{\left(\mathcal{F}_{0,t}+\sum_{j=1}^N L_{j0,t}\Gamma_t^j D_{0j}+\sum_{j=1}^N\sum_{\substack{\ell=1\\\ell\neq j}}^N L_{j\ell,t}\Gamma_t^j D_{j\ell}\right)}_{\mathcal{A}_{\textbf{z},t}}\textbf{z}_t\\
        &\quad+E_0^\top(f_{\mathrm{const},t}-B_t\alpha_t)+\mathcal{K}_{\omega,t+1}\omega_t,
    \end{aligned}
    \label{eqn:z_update}
\end{equation}
where we reuse the already-defined input blocks $B_t^j$ and write

\begin{equation*}
    \begin{aligned}
        \mathcal{F}_{0,t}
        &\coloneqq E_0^\top A_t E_0+\sum_{j=1}^N E_j^\top(I-K_{t+1}^jC_{t+1}^j)A_t E_j,\\
        L_{j0,t}
        &\coloneqq -E_0^\top B_t^j,\qquad
        L_{j\ell,t}
        \coloneqq E_\ell^\top(I-K_{t+1}^\ell C_{t+1}^\ell)B_t^j\quad(\ell\neq j),\\
        \mathcal{K}_{\omega,t+1}
        &\coloneqq\begin{bmatrix}
                I &  &   & \\
                I-K_{t+1}^1C_{t+1}^1 & -K_{t+1}^1 & & \\
                \vdots & & \ddots & \\
                I-K_{t+1}^N C_{t+1}^N & & &  -K_{t+1}^N
            \end{bmatrix}.
    \end{aligned}
\end{equation*}

We define robot $i$'s cost-to-go on the augmented state by

\begin{equation*}
    \mathcal{J}_t^i(\textbf{z}_t)\coloneqq\mathbb{E}\left.\left[\sum_{\tau=t}^T\ell_\tau^i(\textbf{x}_\tau,\textbf{u}_\tau)+\ell_{T+1}^i(\textbf{x}_{T+1})~\right|~\textbf{z}_t\right].
\end{equation*}

\begin{proposition}
\label{prop:quadratic_cost_to_go}
For any sequences $(\Gamma,\alpha)$ and $K$, and for every robot $i\in[N]$ and every stage $t\in[T+1]$, robot $i$'s cost-to-go is quadratic in $\textbf{z}_t$, 

\begin{equation}
        \mathcal{J}_t^i(\textbf{z}_t)=\frac{1}{2}\textbf{z}_t^\top P_t^i\textbf{z}_t+p_t^{i\top}\textbf{z}_t+\frac{1}{2}c_t^i,
        \label{eqn:quadratic_J}
\end{equation}
where $P_t^i, p_t^i, c_t^i$ are computed backward in time, starting from the terminal values

\begin{equation*}
    P_{T+1}^i=E_0^\top Q_{T+1}^iE_0,~ p_{T+1}^i=E_0^\top q_{T+1}^i,~ c_{T+1}^i=2\ell_{\mathrm{const},T+1}^i,
\end{equation*}
and, for $t=T,\dots,1$,

\begin{equation}
        \begin{aligned}
            P_t^i&=E_0^\top Q_t^iE_0+\Gamma_{\textbf{z},t}^\top R_t^i\Gamma_{\textbf{z},t}+\mathcal{A}_{\textbf{z},t}^\top P_{t+1}^i\mathcal{A}_{\textbf{z},t},\\
            p_t^i&=E_0^\top q_t^i+\Gamma_{\textbf{z},t}^\top (R_t^i\alpha_t-r_t^i)
            \\&+\mathcal{A}_{\textbf{z},t}^\top \left[P_{t+1}^iE_0^\top(f_{\mathrm{const},t}-B_t\alpha_t)+p_{t+1}^i\right],\\
            c_t^i&=(R_t^i\alpha_t-2r_t^i)^\top\alpha_t+2\ell_{\mathrm{const},t}^i+c_{t+1}^i+\operatorname{tr}(\mathcal{K}_{\omega,t+1}^\top P_{t+1}^i\mathcal{K}_{\omega,t+1}\Omega_t)\\
            &+\left[(f_{\mathrm{const},t}-B_t\alpha_t)^\top E_0 P_{t+1}^i+2p_{t+1}^{i\top}\right]E_0^\top(f_{\mathrm{const},t}-B_t\alpha_t).
        \end{aligned}
        \label{eqn:Ppc_recursion}
\end{equation}
\end{proposition}
\begin{proof}
    See Appendix~\ref{app:proof_quadratic_cost_to_go}.
\end{proof}

We use the quadratic cost expression in Proposition \ref{prop:quadratic_cost_to_go} to derive first-order necessary conditions for affine-policy stationarity, which eventually leads to the computation of an equilibrium strategy.
To this end, we derive first-order stationarity conditions for the unconditional expected cost $\bar{\mathcal{J}}_t^i=\mathbb{E}[\mathcal{J}_t^i(\textbf{z}_t)]$ with respect to robot $i$'s affine gains $(\Gamma_t^i,\alpha_t^i)$.
We write

\begin{equation}
    \mu_t \coloneqq \mathbb{E}[\textbf{z}_t],\
    \Sigma_{\textbf{z},t}\coloneqq\operatorname{Cov}(\textbf{z}_t),\
    \Xi_t \coloneqq \mathbb{E}[\textbf{z}_t\textbf{z}_t^\top]=\Sigma_{\textbf{z},t}+\mu_t\mu_t^\top.
    \label{eqn:moments_def}
\end{equation}
Then, it follows that
\begin{equation}
    \bar{\mathcal{J}}_t^i
    \coloneqq \mathbb{E}[\mathcal{J}_t^i(\textbf{z}_t)]
    =\frac{1}{2}\operatorname{tr}(P_t^i\Xi_t)+p_t^{i\top}\mu_t+\frac{1}{2}c_t^i
    \label{eqn:expected_augmented_cost}
\end{equation}
is robot $i$'s unconditional expected cost, depending on $(\mu_t,\Xi_t)$.
The moments evolve under~\eqref{eqn:z_update} by

\begin{equation}
    \begin{aligned}
        \mu_{t+1}&=\mathcal{A}_{\textbf{z},t}\mu_t+E_0^\top(f_{\mathrm{const},t}-B_t\alpha_t),\\
        \Sigma_{\textbf{z},t+1}&=\mathcal{A}_{\textbf{z},t}\Sigma_{\textbf{z},t}\mathcal{A}_{\textbf{z},t}^\top+\mathcal{K}_{\omega,t+1}\Omega_t\mathcal{K}_{\omega,t+1}^\top.
    \end{aligned}
    \label{eqn:moment_update}
\end{equation}
With every predictor initialized at the prior mean $\bar{\textbf{x}}_1$ and the stage-1 observation $\textbf{y}_1^i$ update applied, the initial moments are

\begin{equation*}
\mu_1=[\bar{\textbf{x}}_1^\top,\mathbf{0}_{Nn}^\top]^\top,\quad
\Xi_1=\Sigma_{\textbf{z},1}+\mu_1\mu_1^\top,
\end{equation*}
and the blocks of $\operatorname{Cov}(\textbf{z}_1)$ are

\begin{equation*}
\begin{aligned}
\operatorname{Cov}(\textbf{x}_1)&=\Sigma_{\textbf{x},1}^-,\quad \operatorname{Cov}(\textbf{x}_1,\textbf{e}_1^i)=\Sigma_{\textbf{x},1}^-(I-K_1^iC_1^i)^\top,\\
\operatorname{Cov}(\textbf{e}_1^i,\textbf{e}_1^j)
&=(I-K_1^iC_1^i)\Sigma_{\textbf{x},1}^-\cdot(I-K_1^jC_1^j)^\top+\delta_{ij}K_1^iV_1^iK_1^{i\top}.
\end{aligned}
\end{equation*}
Here $\delta_{ij}$ is the Kronecker delta (taking the value $1$ when $i = j$ and $0$ otherwise).
These blocks initialize the forward moment recursion. 
The stagewise first-order conditions are as follows.

\begin{theorem}[First-order Nash conditions for affine strategies]
\label{thm:coupled_expected_stationarity}
Let $(\Gamma,\alpha)$ be a local feedback Nash equilibrium in the affine policy class per Definition \ref{def:local_nash}. Then,

\begin{equation}
\begin{aligned}
\mathcal{M}_t(\mu_t,\Xi_t)
\underbrace{\begin{bmatrix}\operatorname{vec}(\Gamma_t^1)\\\vdots\\\operatorname{vec}(\Gamma_t^N)\\\alpha_t\end{bmatrix}}_{g_t(\Gamma_t,\alpha_t)}
=\nu_t(\mu_t,\Xi_t),\ t\in[T].\\
\end{aligned}
\label{eqn:coupled_gain_alpha_system}
\end{equation}
Here $\mathcal{M}_t(\mu_t,\Xi_t),\nu_t(\mu_t,\Xi_t)$ have the following block structure:
\begin{equation*}
\mathcal{M}_t(\mu_t,\Xi_t)=
\begin{bmatrix}
H_t(\Xi_t)&-\mathfrak{Y}_t(\mu_t)\\
-G_t(\mu_t)&\mathfrak{Z}_t
\end{bmatrix},\quad
\nu_t(\mu_t,\Xi_t)=
\begin{bmatrix}
b_t(\mu_t,\Xi_t)\\
\beta_t(\mu_t)
\end{bmatrix},
\end{equation*}
with blocks partitioned by robot as follows:
\begin{equation*}
\begin{aligned}
H_t(\Xi_t)&=(H_{ik,t})_{i,k=1}^N,&
\mathfrak{Y}_t(\mu_t)&=(\mathfrak{Y}_{ik,t})_{i,k=1}^N,\\
G_t(\mu_t)&=(G_{ik,t})_{i,k=1}^N,&
\mathfrak{Z}_t&=(\mathfrak{Z}_{ik,t})_{i,k=1}^N,\\
b_t(\mu_t,\Xi_t)&=\begin{bmatrix}b_{1,t}^\top & \cdots & b_{N,t}^\top\end{bmatrix}^\top,&
\beta_t(\mu_t)&=\begin{bmatrix}\beta_{1,t}^\top & \cdots & \beta_{N,t}^\top\end{bmatrix}^\top.
\end{aligned}
\end{equation*}
The block-row index $i$ identifies the robot whose stationarity equation is formed. 
The block-column index $k$ identifies the robot whose control gain $\operatorname{vec}(\Gamma_t^k)$ or offset $\alpha_t^k$ multiplies that coefficient block.
The explicit block expressions are given in \eqref{eqn:coupled_blocks} in Appendix~\ref{app:proof_coupled_expected_stationarity}.
Equations~\eqref{eqn:coupled_gain_alpha_system} are first-order stationarity conditions for the affine policies in~\eqref{eqn:actual_single_ctrl}.
A solution defines a local equilibrium under the conditions of  Corollary~\ref{cor:local_nash_affine} that follows.
\end{theorem}
\begin{proof}
    See Appendix~\ref{app:proof_coupled_expected_stationarity}.
\end{proof}

\begin{corollary}[Local Nash equilibrium in the affine class]
\label{cor:local_nash_affine}
Let $(\Gamma^\star,\alpha^\star,K^\star)$ be a candidate affine-policy and estimator triple, and let $(\mu_t^\star,\Xi_t^\star)$ be the mean and second moment of $\textbf{z}_t$ under the closed-loop of that same triple, via~\eqref{eqn:moment_update} from the prior~\eqref{eqn:dgg_initial_cond}.
Suppose $(\Gamma^\star,\alpha^\star)$ solves~\eqref{eqn:coupled_gain_alpha_system} at every stage~$t$ when the linear system is evaluated at $(\mu_t^\star,\Xi_t^\star)$ and at $\{K_{t+1}^{i\star}\}$.
If, at every $(i,t)$, the Hessian of $\bar{\mathcal{J}}_t^i$ with respect to robot~$i$'s own $(\Gamma_t^i,\alpha_t^i)$ is positive definite, then the profile $\gamma^\star$ with
$\gamma_t^{i\star}(\hat{\textbf{x}}_t^i)=-\Gamma_t^{i\star}\hat{\textbf{x}}_t^i-\alpha_t^{i\star}$
is a local feedback Nash equilibrium in the affine class~\eqref{eqn:actual_single_ctrl}.
\end{corollary}
\begin{proof}
    See Appendix~\ref{app:proof_local_nash_affine}.
\end{proof}

\begin{algorithm}[!t]
    \caption{Offline iterative optimization of $(\Gamma,\alpha)$ and $(\mu,\Xi)$}
    \label{alg:nested_moment}
    \begin{algorithmic}[1]
    \STATE \textbf{Initialize:} set $(\Gamma,\alpha)\leftarrow(\Gamma^{\mathrm{init}},\alpha^{\mathrm{init}})$.
    \WHILE{convergence criterion~\eqref{eqn:inner_convergence} and early stopping criterion is not satisfied}
        \STATE \textit{Forward} $(\mu,\Xi)$: initialize from $(\bar{\textbf{x}}_1,\Sigma_{\textbf{x},1}^-)$ and the stage-$1$ measurement update under $K_1$.
        \FOR{$t=1,\dots,T$}
            \STATE Propagate $\mu_{t+1}$, $\Sigma_{\textbf{z},t+1}$ by~\eqref{eqn:moment_update} and form $\Xi_{t+1}$ by~\eqref{eqn:moments_def}.
        \ENDFOR
        \STATE \textit{Backward} $(\Gamma,\alpha)$: freeze $\{(\mu_t,\Xi_t)\}_{t=1}^T$ and $K$; set $P_{T+1}^i=E_0^\top Q_{T+1}^iE_0$, $p_{T+1}^i=E_0^\top q_{T+1}^i$, $c_{T+1}^i=2\ell_{\mathrm{const},T+1}^i$.
        \FOR{$t=T,\dots,1$}
            \STATE Solve~\eqref{eqn:coupled_gain_alpha_system} for $(\Gamma_t,\alpha_t)$ and update the cost-to-go coefficients $(P_t^i,p_t^i,c_t^i)$ using \eqref{eqn:Ppc_recursion}.
        \ENDFOR
    \ENDWHILE
    \RETURN $(\Gamma,\alpha)$ and $\{(\mu_t,\Xi_t)\}$.
    \end{algorithmic}
\end{algorithm}

Note that the stationarity conditions \eqref{eqn:coupled_gain_alpha_system} in Theorem~\ref{thm:coupled_expected_stationarity} rely on the estimator $K$ and the moments $(\mu_t,\Xi_t)$ of the augmented state $\textbf{z}_t$. 
However, the forward propagation of these moments $(\mu_t,\Xi_t)$ depends on the policy parameters $(\Gamma,\alpha)$, which are updated backward in time via dynamic programming.
The opposing recursion directions prevent a direct simultaneous solve of $(\Gamma,\alpha)$ and $(\mu_t,\Xi_t)$.
We propose Algorithm \ref{alg:nested_moment}, which iteratively optimizes $(\Gamma,\alpha)$ (Line 9) and updates $(\mu_t,\Xi_t)$ (Line 5) under fixed $K$.
The optimized $(\Gamma,\alpha)$ then satisfy the stationarity condition \eqref{eqn:coupled_gain_alpha_system} and are 
consistent with the induced moments of the augmented state $\textbf{z}_t$ upon convergence of Algorithm \ref{alg:nested_moment}.

\subsection{Decentralized Sparse Estimation Design}
\label{sec:decentralized_est_gain}

The iterative procedure in Algorithm~\ref{alg:nested_moment} treats the estimator $K$ as given. 
We follow the estimator design approach of Qiu et al. \cite{qiu2026linear} to reduce measurement usage.

We first derive the stepwise minimum-variance dense gain.
Recall the stacked error $\textbf{e}_t=[\textbf{e}_t^{1\top},\dots,\textbf{e}_t^{N\top}]^\top$ and write $\Sigma_t=\operatorname{Cov}(\textbf{e}_t)$.
Stacking~\eqref{eqn:est_update} as in~\cite{qiu2026linear} gives the joint covariance recursion used to design $K$:

\begin{subequations}
    \begin{align}
        \Sigma_{t+1}^-&=(\mathcal{A}_t+\mathcal{B}_t)\Sigma_t(\mathcal{A}_t+\mathcal{B}_t)^\top+\textbf{1}_{N\times N}\otimes W_t,\label{eqn:prior_cov_update}\\
        \Sigma_{t+1}&=(I_{Nn}-\mathcal{K}_{t+1})\Sigma_{t+1}^-(I_{Nn}-\mathcal{K}_{t+1})^\top+\mathbb{K}_{t+1}V_{t+1}\mathbb{K}_{t+1}^\top,\label{eqn:post_cov_update}
    \end{align}
\end{subequations}
with

\begin{equation*}
    \begin{aligned}
        &\mathcal{A}_t=\operatorname{blockdiag}(\{A_t-B_t\Gamma_t\}_{i=1}^N), \mathcal{B}_t=\begin{bmatrix}
                B_t \Pi^1\Gamma_t & \cdots & B_t \Pi^N\Gamma_t\\
                \vdots & \ddots & \vdots \\
                B_t \Pi^1\Gamma_t & \cdots & B_t \Pi^N\Gamma_t
            \end{bmatrix},\\
        &\mathcal{K}_{t+1}=\operatorname{blockdiag}(\{K_{t+1}^i C_{t+1}^i\}_{i=1}^N),\
        \mathbb{K}_{t+1}=\operatorname{blockdiag}(\{K_{t+1}^i\}_{i=1}^N).
    \end{aligned}
    \label{eqn:augmented_system_matrices}
\end{equation*}
Given $\Sigma_t$, the next posterior covariance depends on $\Gamma_t$ and $K_{t+1}$.
Let $\Sigma_t^{i-}\coloneqq\operatorname{Cov}(\textbf{e}_t^{i-})$ denote the $i$-th diagonal $n\times n$ block of the stacked prior $\Sigma_t^-$.
The posterior covariance of robot $i$ is

\begin{equation}
        \Sigma_t^i=(I-K_t^iC_t^i)\Sigma_t^{i-}(I-K_t^iC_t^i)^\top+K_t^iV_t^iK_t^{i\top}.\label{eqn:joseph}
\end{equation}
Setting $\partial\operatorname{tr}(\Sigma_t^i)/\partial K_t^i=\textbf{0}$ yields the stepwise linear minimum-variance (Kalman) gain of the assumed predictor~\cite{welch1995introduction,qiu2026linear}

\begin{equation}
    K_t^{i\star}=\Sigma_t^{i-}C_t^{i\top}(C_t^i\Sigma_t^{i-}C_t^{i\top}+V_t^i)^{-1}.
\label{eqn:alternating_kalman}
\end{equation}

At a fixed stage and prior, each $K_t^{i\star}$ is computed independently.
The prior $\Sigma_t^{i-}$ depends on $\Gamma$ through~\eqref{eqn:prior_cov_update}, and the cost-to-go of Proposition~\ref{prop:quadratic_cost_to_go} depends on $K$ through~\eqref{eqn:z_update}, so the control and estimation designs remain coupled.

\subsubsection{Sparse Group-Lasso Gains}
\label{sec:sparse_estimation}
\label{sec:sparse_est_group_lasso}

The dense gain~\eqref{eqn:alternating_kalman} generally processes every available measurement.
When sensing resources are limited, we instead design a gain $K$ whose sensor blocks can be set to zero.
Algorithm~\ref{alg:nested_moment} computes the affine policy associated with whichever $K$ is given.
Since Proposition~\ref{prop:quadratic_cost_to_go} evaluates expected cost for any $K$, the sparse design can also be compared with the dense Kalman choice to quantify the change in expected cost for using fewer sensors.

We assume that robot~$i$'s observation $\textbf{y}_t^i$ is assembled from $P_i$ sensing groups and can be partitioned as ${\textbf{y}_t^i=[\textbf{y}_t^{i1\top},\dots,\textbf{y}_t^{iP_i\top}]^\top}$.
Correspondingly, we partition the estimation gain into block columns $K_t^i=[K_t^i[1],\cdots,K_t^i[P_i]]$, where $K_t^i[\rho]$  multiplies the innovation associated with sensor~$\rho$. 
In the common inter-robot sensing case, each block corresponds to robot~$i$'s observation of a particular robot~$j$, so sparsity directly determines which robots are observed.

For a fixed prior covariance $\Sigma_t^{i-}$, the unconstrained stepwise minimum-variance gain is given by \eqref{eqn:alternating_kalman}.
This gain generally uses every sensor group.
We therefore seek a sparse approximation that balances fidelity to the Kalman normal equations against a group penalty on the sensor blocks:

\begin{equation}
        \min_{K_t^i}\
        \underbrace{\left\|K_t^i\left(C_t^i\Sigma_t^{i-}C_t^{i\top}+V_t^i\right)
        -\Sigma_t^{i-}C_t^{i\top}\right\|_F^2}_{L_{\mathrm{Estimation}}}
        +\underbrace{\sum_{\rho\in[P_i]}\lambda_t^{i\rho}\left\|K_t^i[\rho]\right\|_F}_{L_{\mathrm{Sparsity}}},
    \label{eqn:lasso_sparse_estimation}
\end{equation}
for all $i\in[N]$ and $t\in[T+1]$.
Here $\|\cdot\|_F$ denotes the Frobenius norm and $\lambda_t^{i\rho}\ge 0$ is the regularization weight for sensor~$\rho$.
The least-squares term penalizes deviations from the dense Kalman normal equations, while the group penalty promotes solutions with fewer active measurement groups.
A larger value of $\lambda_t^{i\rho}$ promotes more aggressive sparsification; smaller values retain more measurements and more closely approximate the dense estimator.
Qiu et al. \cite{qiu2026linear} establish the convexity of the group-lasso problem \eqref{eqn:lasso_sparse_estimation}. We solve it using the fast iterative shrinkage-thresholding algorithm (FISTA) \cite{beck2009fast}.

The continuous group-lasso solution shrinks block norms rather than making a binary sensing decision.
In practice, robot~$i$ ultimately either uses or omits sensor group $\rho$ (omitting $\textbf{y}_t^{i\rho}$ is equivalent to zeroing $K_t^i[\rho]$).
Following \cite{qiu2026linear}, we apply a ratio threshold $r_{\mathrm{th}}\in(0,1]$ and apply the corrective reset

\begin{equation}
K_t^i[\rho] =
\begin{cases}
K_t^{i\star}[\rho], & \|K_t^i[\rho]\|_F\ge r_{\mathrm{th}}\|K_t^{i\star}[\rho]\|_F,
\quad [\textbf{reset}]\\
\mathbf{0}, & \text{otherwise}.\qquad\qquad\quad\  [\textbf{zero-out}]
\end{cases}
\label{eqn:reset_rule}
\end{equation}
Each retained block is restored to its corresponding dense Kalman block (``reset''), and blocks whose norms fall below the threshold are discarded (``zero-out'').
We adopt this hard sparsification to provide on/off sensing decisions while limiting soft attenuation of retained sensors.
A sufficient condition under which the reset is triggered as the estimation-error covariance grows is provided in~\cite{qiu2026linear}.

When sparsity decisions correspond to which robots to observe (inter-robot sensing, where observation blocks are partitioned by robots), the weights $\lambda_t^{ij}$ should reflect how strongly robot~$i$'s control depends on robot~$j$'s state.
That dependence is encoded in the feedback block $\Gamma_t^i[j]$.
Following Qiu et al. \cite{qiu2026linear}, we use the policy-adaptive regularization

\begin{equation}
\lambda_{t+1}^{ij}
:=
\frac{L_1}{\|\Gamma_t^i[j]\|_F+L_2},\quad \lambda_{1}^{ij}
:=
\frac{L_1}{\|\Gamma_1^i[j]\|_F+L_2},
\label{eqn:adaptive_lambda}
\end{equation}
where $L_1,L_2>0$ are tuning parameters.
Large $\|\Gamma_t^i[j]\|_F$ yields a small penalty, so strategically important observations are more likely to be retained.
As the feedback-block norm approaches zero, the penalty approaches $\lambda=\frac{L_1}{L_2}$.
A constant (static) weight $\lambda_t^{ij}\equiv\lambda$ is recovered as the special case in which adaptation is disabled.

In order to avoid numerical instability, we apply sparse regularization only after the dense control-estimation computation has terminated, keeping the resulting feedback gains $(\Gamma,\alpha)$ fixed.

\subsection{Iterative Control–Estimation Optimization}
\label{sec:joint_ilq_alt}
We compute the dense feedback and estimation gains by alternating two steps: update $(\Gamma,\alpha)$ with $K$ fixed using Algorithm~\ref{alg:nested_moment}, then update $K$ for the resulting policy using~\eqref{eqn:alternating_kalman}. The following remark states the conditions satisfied at a fixed point of this iteration.
\begin{remark}
\label{rmk:joint_necessary}
Consider affine strategies of the form $u_t^i=-\Gamma_t^i\hat{\textbf{x}}_t^i-\alpha_t^i$ and linear estimators with gains $\{K_t^i\}$.
Let $(\Gamma^\star,\alpha^\star,K^\star)$ induce, via \eqref{eqn:z_update}-\eqref{eqn:post_cov_update}, the moments $\{(\mu_t^\star,\Xi_t^\star)\}$ and prior covariances $\{\Sigma_t^{i-\star}\}$.
Suppose this triple is a \emph{Nash-Kalman triple} in the following sense:

\begin{enumerate}
    \item \textbf{Control stationarity.}
    For every robot $i$ and stage $t$, the pair
    $(\Gamma_t^{i\star},\alpha_t^{i\star})$ is a first-order stationary point of
    $\bar{\mathcal{J}}_t^i$ with respect to $(\Gamma_t^i,\alpha_t^i)$, given the
    other robots' gains, the estimation gains $\{K^{j\star}\}$, the
    cost-to-go coefficients induced by $(\Gamma^\star,\alpha^\star,\{K^{j\star}\})$,
    and the moments $(\mu_t^\star,\Xi_t^\star)$.
    \item \textbf{Minimum-variance estimation.}
    For every robot $i$ and stage $t$, the gain $K_t^{i\star}$ is a stepwise
    minimum-variance estimator for the prior covariance $\Sigma_t^{i-\star}$
    induced by $(\Gamma^\star,\alpha^\star,\{K^{j\star}\})$.
\end{enumerate}
Then, at every stage $t$,

\begin{enumerate}
    \item[(i)] $(\Gamma_t^\star,\alpha_t^\star)$ solves the coupled system
    \eqref{eqn:coupled_gain_alpha_system} of
    Theorem~\ref{thm:coupled_expected_stationarity} evaluated at
    $(\mu_t^\star,\Xi_t^\star)$ and $\{K_{t+1}^{i\star}\}$;
    \item[(ii)] each $K_t^{i\star}$ satisfies \eqref{eqn:alternating_kalman} evaluated at $\Sigma_t^{i-\star}$.
\end{enumerate}
In particular, any local Nash equilibrium of the affine-strategy game (Corollary~\ref{cor:local_nash_affine}) in which estimators are restricted to the stepwise minimum-variance class must be a fixed point of these two conditions.
\end{remark}

\begin{algorithm}[!t]
    \caption{Offline iterative optimization of $(\Gamma,\alpha)$ and $K$}
    \label{alg:lq_alt_fne}
    \begin{algorithmic}[1]
    \STATE Set $(\Gamma,\alpha)\leftarrow(\Gamma^{\mathrm{init}},\alpha^{\mathrm{init}})$ and $K$ to the corresponding stepwise minimum-variance gains \eqref{eqn:alternating_kalman}.
    \WHILE{convergence criterion~\eqref{eqn:alt_convergence} and early stopping criterion is not satisfied}
        \STATE \textit{Policy Update}: at the current $K$, optimize $(\Gamma,\alpha)$ through Algorithm~\ref{alg:nested_moment}.
        \STATE \textit{Estimator Update}: given the new $(\Gamma,\alpha)$, for $t=1,\dots,T+1$ and each agent $i\in[N]$, $K_t^i\leftarrow\eqref{eqn:alternating_kalman}$.
    \ENDWHILE
    \RETURN $(\Gamma,\alpha,K)$.
    \end{algorithmic}
\end{algorithm}
Again, $K$ is optimized forward in time as the robots' estimation-error covariances propagate forward, while $(\Gamma,\alpha)$ are optimized backward in time.
Following Algorithm~\ref{alg:nested_moment}, we propose the iterative optimization of $(\Gamma,\alpha)$ (Line 3) and $K$ (Line 4) in Algorithm \ref{alg:lq_alt_fne}.
The optimized $(\Gamma,\alpha)$ then satisfy the stationarity condition \eqref{eqn:coupled_gain_alpha_system} and $K$ satisfies the minimum-variance condition \eqref{eqn:alternating_kalman} upon convergence of Algorithm \ref{alg:lq_alt_fne}.

\noindent\textbf{iLQ approximation for nonlinear games.}
The preceding LQ formulation provides a basis for treating nonlinear games. 
Following related literature \cite{fridovich2020efficient,schwarting2021stochastic,laine2023computation}, for general smooth nonlinear costs, dynamics, and observation models as in \eqref{eqn:DTDGG}-\eqref{eqn:ind_obs}, we place the LQ characterization above in an outer iterative linear-quadratic (iLQ) iteration.
At each outer iteration we linearize about a nominal trajectory, run Algorithm~\ref{alg:lq_alt_fne} on the resulting affine-quadratic game, and then roll out a new nominal trajectory using the nonlinear dynamics and estimation update.
In the perfect-state observation setting, fixed points of iLQ algorithms are known to approximately satisfy the first-order conditions associated with feedback Nash equilibria~\cite{fridovich2020efficient,schwarting2021stochastic,laine2023computation}. 

Let $\{\tilde{\textbf{x}}_t\}_{t=1}^{T+1}$ and $\{\tilde{\textbf{u}}_t\}_{t=1}^{T}$ be nominal state and control trajectories satisfying $\tilde{\textbf{x}}_{t+1}=f_t(\tilde{\textbf{x}}_t,\tilde{\textbf{u}}_t)$. We approximate the costs, dynamics, and observations about these trajectories by the following affine-quadratic model:

\begin{equation*}
    \begin{aligned}
        \ell_t^i(\textbf{x}_t,\textbf{u}_t)&\approx\frac{1}{2}\textbf{x}_t^\top Q_t^i\textbf{x}_t+q_t^{i\top}\textbf{x}_t+\frac{1}{2}\textbf{u}_t^\top R_t^i\textbf{u}_t+r_t^{i\top}\textbf{u}_t
        \\
        &+ \underbrace{\ell_t^i(\Tilde{\textbf{x}}_t,\Tilde{\textbf{u}}_t)+\frac{1}{2}\Tilde{\textbf{x}}_t^\top Q_t^i\Tilde{\textbf{x}}_t-\ell_{\textbf{x},t}^{i\top}\Tilde{\textbf{x}}_t+\frac{1}{2}\Tilde{\textbf{u}}_t^\top R_t^i\Tilde{\textbf{u}}_t-\ell_{\textbf{u},t}^{i\top}\Tilde{\textbf{u}}_t}_{\ell_{\mathrm{const},t}^i},\\
        f_t(\textbf{x}_t,\textbf{u}_t)&\approx A_t\textbf{x}_t+B_t\textbf{u}_t+\underbrace{f_t(\Tilde{\textbf{x}}_t,\Tilde{\textbf{u}}_t)-A_t\Tilde{\textbf{x}}_t-B_t\Tilde{\textbf{u}}_t}_{f_{\text{const},t}},\\
        h_t^i(\textbf{x}_t)&\approx C_t^i\textbf{x}_t+\underbrace{h_t^i(\Tilde{\textbf{x}}_t)-C_t^i\Tilde{\textbf{x}}_t}_{h_{\text{const},t}^i},\\
    \end{aligned}
\end{equation*}
where

\begin{equation*}
    \begin{aligned}
        &Q_t^i\coloneqq\ell_{\textbf{xx},t}^i,
        \ q_t^i\coloneqq\ell_{\textbf{x},t}^i-\ell_{\textbf{xx},t}^i\Tilde{\textbf{x}}_t,\ R_t^i\coloneqq\ell_{\textbf{uu},t}^i,
        \ r_t^i\coloneqq\ell_{\textbf{u},t}^i-\ell_{\textbf{uu},t}^i\Tilde{\textbf{u}}_t,\\
        &A_t\coloneqq f_{\textbf{x},t}^\top,
        \ B_t\coloneqq f_{\textbf{u},t}^\top,\qquad
        C_t^i\coloneqq h_{\textbf{x},t}^{i\top},
    \end{aligned}
\end{equation*}
and $\ell_{\textbf{xx},t}^i,\ell_{\textbf{x},t}^i,\ell_{\textbf{uu},t}^i,\ell_{\textbf{u},t}^i,f_{\textbf{x},t},f_{\textbf{u},t},h_{\textbf{x},t}^i$ denote the Hessians and gradients of $\ell_t^i$, $f_t$, and $h_t^i$ evaluated at the nominal state $\Tilde{\textbf{x}}_t$ and control $\Tilde{\textbf{u}}_t$.

Algorithm~\ref{alg:ilq_alt_fne} summarizes the combination of iterative optimization and iLQ approximation for nonlinear games.
From the task reference $(\tilde{\textbf{x}}^{\mathrm{ref}},\tilde{\textbf{u}}^{\mathrm{ref}})$, a deterministic iLQ Nash solve produces $(\tilde{\textbf{x}}^{\mathrm{det}},\tilde{\textbf{u}}^{\mathrm{det}},\Gamma^{\mathrm{det}},\alpha^{\mathrm{det}})$.
The outer iteration then linearizes about the current nominal trajectory, runs Algorithm~\ref{alg:lq_alt_fne} on that local affine-quadratic game, and rolls out a new nominal trajectory on the nonlinear dynamics.

\begin{algorithm}[!t]
    \caption{Iterative LQ approximation for Algorithm \ref{alg:lq_alt_fne} in nonlinear games}
    \label{alg:ilq_alt_fne}
    \begin{algorithmic}[1]
    \STATE \textbf{Deterministic iLQ Nash Initialization:} using $(\tilde{\textbf{x}}^{\mathrm{ref}},\tilde{\textbf{u}}^{\mathrm{ref}})$, solve the nonlinear game by iterative LQ approximation to obtain $(\tilde{\textbf{x}}^{\mathrm{det}},\tilde{\textbf{u}}^{\mathrm{det}},\Gamma^{\mathrm{det}},\alpha^{\mathrm{det}})$.
    \STATE Set $(\tilde{\textbf{x}},\tilde{\textbf{u}},\Gamma,\alpha)\leftarrow(\tilde{\textbf{x}}^{\mathrm{det}},\tilde{\textbf{u}}^{\mathrm{det}},\Gamma^{\mathrm{det}},\alpha^{\mathrm{det}})$.
    \WHILE{convergence criterion~\eqref{eqn:ilq_convergence} and early stopping criterion is not satisfied}
        \STATE \textit{Linearization}: Based on $(\tilde{\textbf{x}},\tilde{\textbf{u}})$, form the LQ data $\{Q_t^i,q_t^i,R_t^i,r_t^i,A_t,B_t,C_t^i,f_{\mathrm{const},t},h_{\mathrm{const},t}^i\}$ and the terminal quadratic $\{Q_{T+1}^i,q_{T+1}^i\}$.
        \STATE \textit{Policy-Estimator Update}: run Algorithm~\ref{alg:lq_alt_fne} from $(\Gamma,\alpha)$ to obtain $(\Gamma,\alpha,K)$.
        \STATE \textit{Rollout}: apply $u_t^i=-\Gamma_t^i\hat{\textbf{x}}_t-\alpha_t^i$ on the nonlinear plant from the prior $\hat{\textbf{x}}_t^-$ to obtain a new nominal $(\tilde{\textbf{x}},\tilde{\textbf{u}})$.
    \ENDWHILE
    \RETURN $(\Gamma,\alpha,K)$.
    \end{algorithmic}
\end{algorithm}

\noindent\textbf{Convergence criteria.}
For the numerical iterations, we set $\epsilon_{\mathrm{ALT}}=\epsilon_{\mathrm{iLQ}}=10^{-2}$ as the stopping tolerances for the control-estimation and outer iLQ iterations.
The successive-iterate residuals are

\begin{equation*}
\begin{aligned}
d_\Gamma&\coloneqq\frac{1}{NT}\sum_{t=1}^{T}
    \|\Gamma_t^{\mathrm{new}}-\Gamma_t^{\mathrm{old}}\|_F,\quad d_\alpha\coloneqq\frac{1}{NT}\sum_{t=1}^{T}
    \|\alpha_t^{\mathrm{new}}-\alpha_t^{\mathrm{old}}\|_2,\\
d_K&\coloneqq\frac{1}{N(T+1)}\sum_{t=1}^{T+1}
    \sqrt{\sum_{i=1}^{N}\|K_t^{i,\mathrm{new}}-K_t^{i,\mathrm{old}}\|_F^2},
\end{aligned}
\end{equation*}
and, for the outer iLQ iteration,

\begin{equation*}
    d_x\coloneqq\frac{1}{N(T+1)}\sum_{t=1}^{T+1}\|\tilde{\textbf{x}}_t^{\mathrm{new}}-\tilde{\textbf{x}}_t^{\mathrm{old}}\|_2,
    \ 
    d_u\coloneqq\frac{1}{NT}\sum_{t=1}^{T}\|\tilde{\textbf{u}}_t^{\mathrm{new}}-\tilde{\textbf{u}}_t^{\mathrm{old}}\|_2.
\end{equation*}
Using the successive-iterate changes above, we consider the inner policy-moment iteration converged when

\begin{equation}
    \max\{d_\Gamma,d_\alpha\}<\epsilon_{\mathrm{ALT}},
    \label{eqn:inner_convergence}
\end{equation}
the control-estimation iteration converged when

\begin{equation}
    \max\{d_\Gamma,d_\alpha,d_K\}<\epsilon_{\mathrm{ALT}},
    \label{eqn:alt_convergence}
\end{equation}
and the outer iLQ iteration converged when

\begin{equation}
    \max\{d_x,d_u\}<\epsilon_{\mathrm{iLQ}}.
    \label{eqn:ilq_convergence}
\end{equation}

\noindent\textbf{Early-stopping criteria.}
In the inner iteration of Algorithm \ref{alg:nested_moment}, if either $d_\Gamma$ or $d_\alpha$ increases relative to the preceding step, we reject the candidate and return the last accepted triple $(\Gamma,\alpha,K)$.
In the control-estimation iteration (Algorithm \ref{alg:lq_alt_fne}), we instead evaluate the mean coupled-control stationarity residual defined as follows:
\begin{equation}
    \mathcal R
    \coloneqq\frac{1}{NT}\sum_{t=1}^{T}\sum_{i=1}^{N}
    \left\|
    \left[\mathcal M_t(\mu_t,\Xi_t)g_t(\Gamma_t,\alpha_t)
    -\nu_t(\mu_t,\Xi_t)\right]_i
    \right\|_2.
    \label{eqn:dense_stationarity_residual}
\end{equation}
Here $[\cdot]_i$ selects the rows associated with robot $i$'s feedback gain $\Gamma_t^i$ and affine term $\alpha_t^i$.
A candidate is rejected if $\mathcal{R}$ increases.

In Section \ref{sec:experiment}, we will show that the convergence and early-stopping criteria are effective for $(\Gamma,\alpha)$ to satisfy the stationarity condition of \eqref{eqn:coupled_gain_alpha_system} across robot tasks.

\noindent\textbf{Online deployment.}
Algorithm \ref{alg:online_eval} summarizes online deployment of the computed feedback policy and estimator.

\begin{algorithm}[!t]
    \caption{Online deployment of the policy and estimator}
    \label{alg:online_eval}
    \begin{algorithmic}[1]
    \FOR{$t=1,\dots,T$}
        \FOR{$i=1,\dots,N$}
            \STATE \textit{Estimation}: $\hat{\textbf{x}}_t^i\leftarrow\hat{\textbf{x}}_t^{i-}+K_t^i\left[\textbf{y}_t^i-h_t^i(\hat{\textbf{x}}_t^{i-})\right]$.
            \STATE \textit{Control Computation}: $u_t^i\leftarrow-\Gamma_t^i\hat{\textbf{x}}_t^i-\alpha_t^i$.
            \STATE \textit{State Prediction}: $\hat{\textbf{x}}_{t+1}^{i-}\leftarrow f_t(\hat{\textbf{x}}_t^i,\hat{\textbf{u}}_t^i)$.
        \ENDFOR
        \STATE \textit{State Update}: $\textbf{x}_{t+1}\leftarrow f_t(\textbf{x}_t,\textbf{u}_t)+\textbf{w}_t$.
    \ENDFOR
    \FOR{$i=1,\dots,N$}
        \STATE 
        \textit{Estimation}: ${\hat{\textbf{x}}_{T+1}^i\leftarrow\hat{\textbf{x}}_{T+1}^{i-}+K_{T+1}^i\left[\textbf{y}_{T+1}^i-h_{T+1}^i(\hat{\textbf{x}}_{T+1}^{i-})\right]}$.
    \ENDFOR
    \end{algorithmic}
\end{algorithm}

In summary, we propose an iterative control-estimation optimization framework for the stochastic game with individual observations that computes a feedback Nash strategy $(\Gamma,\alpha)$ and a minimum-variance estimator $K$.
We evaluate the optimized $(\Gamma,\alpha,K)$ across various multi-robot tasks in Section \ref{sec:experiment}.

\section{Experiments}
\label{sec:experiment}
\subsection{Evaluation Setup}
\begin{figure*}[t]
\centering
\includegraphics[width=\linewidth]{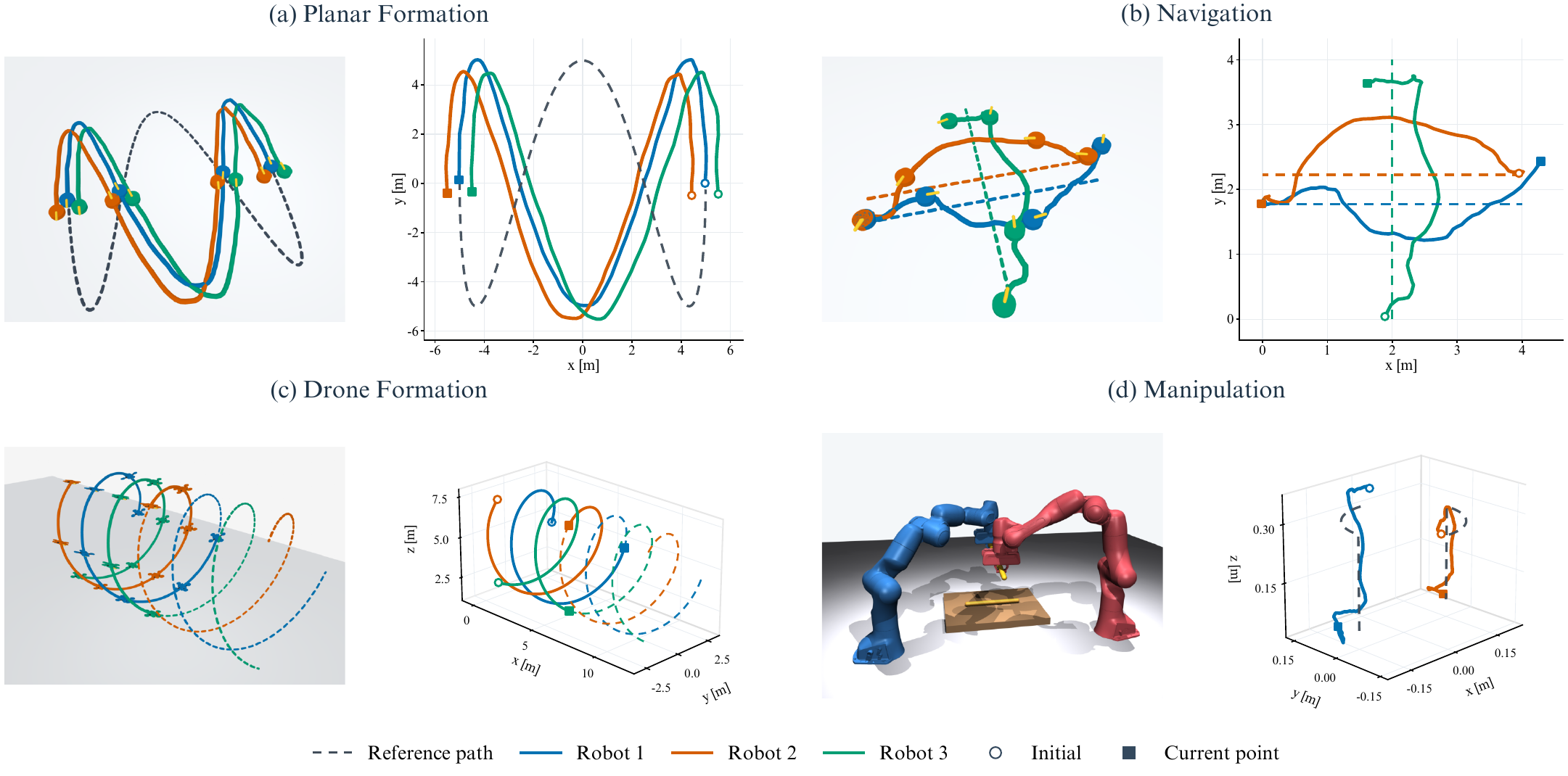}
\caption{Illustrations of the four multi-robot tasks: planar formation, navigation, drone formation, and manipulation.
The left image in each panel shows a snapshot from the MuJoCo\cite{todorov2012mujoco} simulator and the right image plots the robot trajectories.}
\label{fig:task_illustrations}
\end{figure*}

We evaluate the method from Sec.~\ref{sec:method} on four example multi-robot tasks: planar formation, navigation, drone formation, and manipulation (Fig.~\ref{fig:task_illustrations}), using dense and sparse estimators. 
In dense estimator experiments, we evaluate the joint control-estimation design proposed in Algorithm \ref{alg:lq_alt_fne} (and Algorithm \ref{alg:ilq_alt_fne} for settings with nonlinear dynamics/observations, and/or nonquadratic objectives). In sparse estimator experiments, we evaluate robot performance using the policy-adaptive sparse estimator optimized through \eqref{eqn:lasso_sparse_estimation}.
For each task, we conduct Monte Carlo simulations with 100 trials, where the random seeds are the same across methods.
We also evaluate our proposed method in robot hardware validation experiments to validate its capability of real-time control and estimation of multi-robot systems.

\subsection{Simulation Tasks}
\subsubsection{Planar Formation}
\label{sec:experiment_planar}
The planar formation task is inspired by the NASA CADRE mission~\cite{saboia2024cadre} discussed in Sec.~\ref{sec:introduction}.
A leading robot tracks a reference path and two following robots maintain desired displacements relative to the leader and one another.

\noindent\textbf{Dynamics and cost.}
We evaluate \textit{linear} double-integrator (DI) and \textit{nonlinear} unicycle (UC) dynamical models with sampling interval $\Delta t=\SI{0.1}{\second}$. Let $s_{p,t}^i=[s_{p,x,t}^i,s_{p,y,t}^i]^\top$ be position and $s_{v,t}^i=[s_{v,x,t}^i,s_{v,y,t}^i]^\top$ be Cartesian velocity in the DI model. The robot state is $x_t^i=[(s_{p,t}^i)^\top,(s_{v,t}^i)^\top]^\top$, and the control $u_t^i=[a_{x,t}^i,a_{y,t}^i]^\top$ is Cartesian acceleration. Its dynamics are

\begin{equation}
\begin{aligned}
s_{p,t+1}^i&=s_{p,t}^i+\Delta t\,s_{v,t}^i+\tfrac12\Delta t^2u_t^i+w_{p,t}^i,\\
s_{v,t+1}^i&=s_{v,t}^i+\Delta t\,u_t^i+w_{v,t}^i.
\end{aligned}\tag{DI-Dynamics}
\end{equation}
In the UC model, the robot state $x_t^i=[(s_{p,t}^i)^\top,\psi_t^i,s_t^i]^\top$ contains position, heading angle $\psi_t^i$, and forward speed $s_t^i$. The control $u_t^i=[\zeta_t^i,a_t^i]^\top$ contains heading rate $\zeta_t^i$ and forward acceleration $a_t^i$. The dynamics are

\begin{equation}
\begin{aligned}
s_{p,t+1}^i&=s_{p,t}^i+\Delta t\,s_t^i\begin{bmatrix}
    \cos\psi_t^i & \sin\psi_t^i
\end{bmatrix}^\top+w_{p,t}^i,\\
\psi_{t+1}^i&=\psi_t^i+\Delta t\,\zeta_t^i+w_{\psi,t}^i,\\
s_{t+1}^i&=s_t^i+\Delta t\,a_t^i+w_{s,t}^i.
\end{aligned}\tag{UC-Dynamics}
\end{equation}
The reference path for Robot 1 (R1) and desired displacements between robots are

\begin{equation*}
\begin{aligned}
s_{p,t,\text{ref}}&=5\begin{bmatrix}
    \cos((t-1)\Delta t/3) & \sin((t-1)\Delta t)
\end{bmatrix}^\top,\\
d^{21}&=\begin{bmatrix}
    -0.5\\-0.5
\end{bmatrix},\  d^{31}=\begin{bmatrix}
    0.5\\-0.5
\end{bmatrix},\
d^{23}=\begin{bmatrix}
    -1\\0
\end{bmatrix},\ d^{ji}=-d^{ij}.
\end{aligned}
\end{equation*}
Each robot's running cost and terminal cost are \textit{quadratic}:

\begin{equation*}
\begin{aligned}
\ell_t^1&=\kappa_t^{1\top}\cdot\begin{bmatrix}
    \|s_{p,t}^1-s_{p,t,\text{ref}}\|_2^2 & \|u_t^1\|_2^2
\end{bmatrix}^\top,\\
\ell_{T+1}^1&=\kappa_{T+1}^1\cdot\|s_{p,T+1}^1-s_{p,T+1,\text{ref}}\|_2^2,\\
\ell_t^i&=\kappa_t^{i\top}\cdot\begin{bmatrix}
    \sum_{j\ne i}\|s_{p,t}^i-s_{p,t}^j-d^{ij}\|_2^2 & \|u_t^i\|_2^2
\end{bmatrix}^\top,\\
\ell_{T+1}^i&=\kappa_{T+1}^i\cdot \sum_{j\ne i}\|s_{p,T+1}^i-s_{p,T+1}^j-d^{ij}\|_2^2,
\quad i\in\{2,3\}.
\end{aligned}
\end{equation*}
The vectors $\kappa_t^i$ contain the running-cost weights; terminal weights are denoted by $\kappa_{T+1}^i$ for all tasks and are listed in Table~\ref{tab:agent_cost_weights}.

\noindent\textbf{Observation models.}
We use two observation models: a \textit{linear} identity sensing model (I-Obs) and a \textit{nonlinear} decentralized range-bearing sensing model (D-Obs). Under I-Obs, every robot receives an independently corrupted copy of the joint state:

\begin{equation}
\textbf{y}_t^i=\textbf{x}_t+v_t^i.\tag{I-Obs.}
\end{equation}
Under D-Obs, robot $i$ receives one four-channel block per robot. Set $\Delta s_{p,t}^{ij}=s_{p,t}^j-s_{p,t}^i$ for $j\ne i$ and $\Delta s_{p,t}^{i0}=-s_{p,t}^i$ for the self block's link to the fixed origin. The range and global bearing of either link are

\begin{equation*}
\begin{aligned}
\rho_t^{ij}=\|\Delta s_{p,t}^{ij}\|_2,\ \varphi_t^{ij}=\operatorname{atan2}(\Delta s_{p,y,t}^{ij},\Delta s_{p,x,t}^{ij}),
\ j\in\{0,\dots,3\}\setminus\{i\}.
\end{aligned}
\end{equation*}
With $h_{t,\text{DI}}^{ij}$ denoting the block associated with target $j$, the D-Obs model for DI dynamics is

\begin{subequations}
\begin{align}
\textbf{y}_t^i&=h_{t,\text{DI}}^i(\textbf{x}_t)+v_t^i,\notag\\
h_{t,\text{DI}}^{ii}&=[\rho_t^{i0},\varphi_t^{i0},s_{v,x,t}^i,s_{v,y,t}^i]^\top,\tag{Self Obs.}\\
h_{t,\text{DI}}^{ij}&=[\rho_t^{ij},\varphi_t^{ij},s_{v,x,t}^j,s_{v,y,t}^j]^\top,\quad j\ne i.\tag{Inter-robot Obs.}
\end{align}
\end{subequations}
For UC dynamics, the observation model is

\begin{subequations}
\begin{align}
\varphi_{\text{body},t}^{ij}&=\operatorname{wrap}(\varphi_t^{ij}-\psi_t^i),\notag\\
\textbf{y}_t^i&=h_{t,\text{UC}}^i(\textbf{x}_t)+v_t^i,\notag\\
h_{t,\text{UC}}^{ii}&=[\rho_t^{i0},\varphi_{\text{body},t}^{i0},\psi_t^i,s_t^i]^\top,\tag{Self Obs.}\\
h_{t,\text{UC}}^{ij}&=[\rho_t^{ij},\varphi_{\text{body},t}^{ij},\psi_t^j,s_t^j]^\top,\quad j\ne i.\tag{Inter-robot Obs.}
\end{align}
\end{subequations}

\noindent\textbf{Noise scales with distance.} In practice, observation uncertainty often increases with sensing distance.
We therefore introduce the following noise covariance scaling pattern for D-Obs:

\begin{equation*}
    {\sigma_t^{ij}}^2=\sigma_0^2+\left(a_{\text{noise}}\cdot\|s_{p,t}^i-s_{p,t}^j\|_2\right)^2,
\end{equation*}
where $a_{\text{noise}}$ determines how quickly observation noise grows with distance $\|s_{p,t}^i-s_{p,t}^j\|_2$ between robot $i$ and robot $j$. 
For I-Obs, the standard deviation of process noise and observation noise remain constant as $\sigma_0$.

\subsubsection{Navigation}
\label{sec:experiment_navigation}
Three robots track intersecting reference paths while penalizing proximity. 
Navigation tasks use the same DI and UC dynamics and the same I-Obs and D-Obs models as planar formation. 
With $s_{p,t,\text{ref}}^i$ denoting robot $i$'s reference, its \textbf{nonlinear} running cost and terminal cost are

\begin{equation*}
\begin{aligned}
\ell_t^i&=\kappa_t^{i\top}\cdot\begin{bmatrix}
    \|s_{p,t}^i-s_{p,t,\text{ref}}^i\|_2^2 & \sum_{j\ne i}\frac{1}{\|s_{p,t}^i-s_{p,t}^j\|_2^2+\delta} & \|u_t^i\|_2^2
\end{bmatrix}^\top,\\
\ell_{T+1}^i&=\kappa_{T+1}^{i\top}\begin{bmatrix}
    \|s_{p,T+1}^i-s_{p,T+1,\text{ref}}^i\|_2^2 & \sum_{j\ne i}\frac{1}{\|s_{p,T+1}^i-s_{p,T+1}^j\|_2^2+\delta}
\end{bmatrix}^\top.
\end{aligned}
\end{equation*}

\subsubsection{Drone Formation}
\label{sec:experiment_drone}
Three identical quadrotors follow separate strands of a triple helix and form a rotating triangle as they advance through 3-D space.
Each drone tracks its assigned path while maintaining the desired relative displacements to the other two drones.

\noindent\textbf{Dynamics and cost.}
For each drone $i$, let

\begin{equation*}
    \begin{aligned}
        x_t^i&=\begin{bmatrix}(s_{p,t}^i)^\top&(\eta_t^i)^\top&(s_{v,t}^i)^\top&(\varpi_t^i)^\top\end{bmatrix}^\top\in\mathbb R^{12},
    \\u_t^i&=\begin{bmatrix}u_{1,t}^i&u_{2,t}^i&u_{3,t}^i&u_{4,t}^i\end{bmatrix}^\top,
    \end{aligned}
\end{equation*}
where $s_{p,t}^i$ and $s_{v,t}^i$ are world-frame position and velocity,
$\eta_t^i=[\phi_t^i,\theta_t^i,\psi_t^i]^\top$ contains the ZYX Euler angles, $\varpi_t^i$ is body angular
velocity, and $u_{k,t}^i$ is rotor $k$'s thrust deviation from hover. We use rigid-body quadrotor dynamics in ZYX Euler coordinates, with thrust along the body $z$ axis and rotor-generated body torques~\cite{busetto2026nonlinear}:

\begin{equation*}
\begin{aligned}
    \dot s_{p,t}^i &= s_{v,t}^i, &
    \dot s_{v,t}^i &= -g e_3+\frac{m_qg+\mathbf 1^\top u_t^i}{m_q}\mathsf R(\eta_t^i)e_3, \\
    \dot\eta_t^i &= \mathsf T_{\text{Euler}}(\eta_t^i)\varpi_t^i,&
    \dot\varpi_t^i &= \mathsf I_{\text{body}}^{-1}\!\left(\tau(u_t^i)-\varpi_t^i\times \mathsf I_{\text{body}}\varpi_t^i\right).
\end{aligned}
\end{equation*}
The matrix $\mathsf I_{\text{body}}$ is the inertia tensor about the center of mass expressed in body coordinates; $\mathsf I_{\text{body}}^{-1}$ denotes its matrix inverse.
$e_3=[0,0,1]^\top$ is the upward unit vector in the world frame, $\mathbf{1}\in\mathbb R^4$ is the all-ones vector, $\mathsf R(\eta_t^i)$ is the ZYX body-to-world rotation, and $\mathsf T_{\text{Euler}}(\eta_t^i)$ maps body rates
to Euler-angle rates. The torque mixing map is

\begin{equation*}
\tau(u_t^i)=\begin{bmatrix}
l\cdot(u_{2,t}^i-u_{4,t}^i)\\
l\cdot(u_{3,t}^i-u_{1,t}^i)\\
\zeta\cdot(u_{1,t}^i-u_{2,t}^i+u_{3,t}^i-u_{4,t}^i)
\end{bmatrix},
\end{equation*}
with arm length $l$ and yaw torque-thrust ratio $\zeta$.
We use $m_q=\SI{1}{\kilogram}$, $g=\SI{9.81}{\metre/\second\squared}, l=\SI{0.20}{\metre}, \zeta=\SI{0.02}{\metre}$ and ${\mathsf I_{\text{body}}=\operatorname{diag}(0.005,0.005,0.009)\,\si{\kilogram\cdot\metre\squared}}$.
We integrate the continuous dynamics using fourth-order Runge–Kutta (RK4) integration following ~\cite{busetto2026nonlinear}, and add process noise after the step:

\begin{equation*}
    x_{t+1}^i=\operatorname{RK4}_{\Delta t}(x_t^i,u_t^i)+w_t^i,
    \qquad \Delta t=\SI{0.05}{\second}.
\end{equation*}
The reference path is a $+x$-oriented triple helix that lasts $\SI{12}{\second}$ ($T=240$), advances
at $\SI{1}{\metre/\second}$, has a radius of $\SI{3}{\metre}$ and a center altitude of
$\SI{4.5}{\metre}$, and completes two turns. With phase
${\vartheta_t=(2\pi/6)(t-1)\Delta t}$ and $\delta_i=2\pi(i-1)/3$, each robot tracks its own strand for $t\in[T+1]$:

\begin{equation*}
    s_{p,t,\text{ref}}^i=
    \begin{bmatrix}
        (t-1)\Delta t & 3\cos(\vartheta_t+\delta_i) &
        4.5+3\sin(\vartheta_t+\delta_i)
    \end{bmatrix}^\top.
\end{equation*}
Each drone also maintains the desired relative displacements to the other two drones:

\begin{equation*}
\begin{aligned}
    \ell_t^i&=\kappa_t^{i\top}\cdot\begin{bmatrix}
        \|s_{p,t}^i-s_{p,t,\text{ref}}^i\|_2^2 \\ \sum_{j\ne i}\|s_{p,t}^i-s_{p,t}^j-(s_{p,t,\text{ref}}^i-s_{p,t,\text{ref}}^j)\|_2^2 \\ \|u_t^i\|_2^2
    \end{bmatrix},\\
    \ell_{T+1}^i&=\kappa_{T+1}^{i\top}\cdot\begin{bmatrix}
    \|s_{p,T+1}^i-s_{p,T+1,\text{ref}}^i\|_2^2 \\ \sum_{j\ne i}\|s_{p,T+1}^i-s_{p,T+1}^j-(s_{p,T+1,\text{ref}}^i-s_{p,T+1,\text{ref}}^j)\|_2^2
\end{bmatrix}.
\end{aligned}
\end{equation*}

\noindent\textbf{Observation model.}
Each drone observes an independently corrupted copy of the 36-dimensional joint state under the identity (I-Obs) model.

\subsubsection{Manipulation}
\label{sec:experiment_arm}
Two Franka Panda robots, each with seven actuated degrees of freedom, face each other in the MuJoCo \cite{todorov2012mujoco} simulator.
They grasp opposite ends of a rigid rod and coordinate a $90^\circ$ rotation about the vertical axis, moving the rod from alignment with $+y$ to alignment with $+x$.
Each robot tracks its assigned end-effector pose while maintaining the desired displacement between the grippers under individual noisy observations.

\noindent\textbf{Dynamics and cost.}
Each robot follows nonlinear joint dynamics. For robot $i$, $\xi_t^i\in\mathbb R^7$ is the vector of actuated joint angles, $\dot \xi_t^i\in\mathbb R^7$ is the joint-velocity vector, and $u_t^i\in\mathbb R^7$ is the commanded joint-torque vector. Thus ${x_t^i=[(\xi_t^i)^\top,(\dot \xi_t^i)^\top]^\top\in\mathbb R^{14}}$ evolves as follows:
\begin{equation*}
    x_{t+1}^i=f_{\text{MJ}}\big(x_t^i,u_t^i+\tau_{\text{bias}}(x_t^i)\big)+w_t^i,
    \qquad \Delta t=\SI{0.05}{\second},
\end{equation*}
where $f_{\text{MJ}}$ denotes one nonlinear MuJoCo step and $\tau_{\text{bias}}$ compensates the simulated joint bias forces. The iLQ solver uses a finite-difference linearization of this discrete-time map about each nominal state and control.
The horizon is $\SI{4}{\second}$ ($T=80$). 

Let $s_{p,t}^i=s_p^i(\xi_t^i)$ and $\mathsf R_t^i=\mathsf R^i(\xi_t^i)$ be the end-effector position and orientation obtained from forward kinematics. 
The reference position $s_{p,t,\text{ref}}^i$ is the desired rod-end position, $s_{p,t}^i$ is the end-effector position, $\mathsf R_{t,\text{des}}^i$ is the desired palm-down grasp orientation, and $e_{\text{rot}}(\mathsf R_t^i,\mathsf R_{t,\text{des}}^i)$ is their rotation-vector error. For $j\ne i$, define $\Delta s_{p,t,\text{ref}}^{ij}=s_{p,t,\text{ref}}^j-s_{p,t,\text{ref}}^i$.
Arm $i$'s running and terminal costs are

\begin{equation*}
\begin{aligned}
    \ell_t^i&=\kappa_t^{i\top}\cdot\begin{bmatrix}
        \|s_{p,t}^i-s_{p,t,\text{ref}}^i\|_2^2\\
        \|e_{\text{rot}}(\mathsf R_t^i,\mathsf R_{t,\text{des}}^i)\|_2^2\\
        \left\|(s_{p,t}^j-s_{p,t}^i)-\Delta s_{p,t,\text{ref}}^{ij}\right\|_2^2\\
        \|u_t^i\|_2^2
    \end{bmatrix},\\
    \ell_{T+1}^i&=\kappa_{T+1}^{i\top}\cdot\begin{bmatrix}
    \|s_{p,T+1}^i-s_{p,T+1,\text{ref}}^i\|_2^2\\
    \|e_{\text{rot}}(\mathsf R_{T+1}^i,\mathsf R_{T+1,\text{des}}^i)\|_2^2\\
    \left\|(s_{p,T+1}^j-s_{p,T+1}^i)-\Delta s_{p,T+1,\text{ref}}^{ij}\right\|_2^2
\end{bmatrix}.
\end{aligned}
\end{equation*}

\noindent\textbf{Observation model.}
Each arm observes an independently corrupted copy of the 28-dimensional joint state under the identity (I-Obs) model.

\begin{figure*}[t]
\centering
\includegraphics[width=\linewidth]{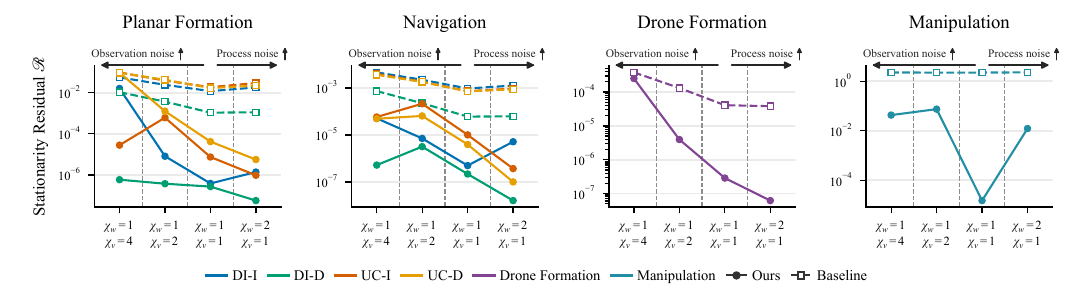}
\caption{Stationarity residual $\mathcal R$ across noise levels, dynamics, and observation models (solid curves: our proposed method; dashed curves: baseline).
}
\label{fig:dense_stationarity_noise}
\end{figure*}

\subsection{Evaluation Metrics}
\label{sec:eval_metrics}
\subsubsection{Robot Cost}
We compute analytical expected costs from \eqref{eqn:expected_augmented_cost} and evaluate realized performance using 100 Monte Carlo rollouts with realized states and controls.\footnote{Robot cost is an empirical performance metric for these general-sum examples; it does not measure equilibrium quality or benefit to every robot.}
We normalize the total cost by the number of time-steps $T+1$, and report both costs per robot in the task and the average across robots.
For games with nonlinear dynamics, observations, or costs, analytical costs are evaluated on the corresponding iLQ approximation, while realized costs use nonlinear rollouts and the original costs.

\subsubsection{Stationarity Residual}
We measure how closely each each computed policy and its associated estimators satisfy the coupled affine-control stationarity equation~\eqref{eqn:coupled_gain_alpha_system}, using \eqref{eqn:dense_stationarity_residual}. 
Each policy uses its own saved estimation gains, induced moments, and backward cost-to-go coefficients. 
For nonlinear tasks, we evaluate the local LQ model formed at the nominal trajectory used in the iLQ iteration that produced its reported control gains, including the deterministic baseline's control solve. 
We propagate $\mu_t$ and $\Xi_t$ analytically on that fixed model using the specified noise covariances.

\subsubsection{Task-specific Metrics}
For planar formation, drone formation, and manipulation rollouts, each robot's task metric is the root-mean-square error:

\begin{equation*}
\text{RMSE}^i=\left(\frac{1}{T+1}\sum_{t=1}^{T+1}\varepsilon_{t,i}^2\right)^{1/2},
\end{equation*}
\begin{subequations}
\begin{align}
\varepsilon_{t,1}^2&=\|s_{p,t}^1-s_{p,t,\text{ref}}\|_2^2,\tag{Planar formation -- \textbf{Leader}}\\
\varepsilon_{t,i}^2&=\frac{1}{N-1}\sum_{j\ne i}\|s_{p,t}^i-s_{p,t}^j-d^{ij}\|_2^2,\tag{Planar formation -- \textbf{Follower}}\\
\varepsilon_{t,i}^2&=\|s_{p,t}^i-s_{p,t,\text{ref}}^i\|_2^2. \tag{Drone formation and manipulation}
\end{align}
\end{subequations}
Navigation instead uses the minimum pairwise separation over the rollout,

\begin{equation*}
d_{\min}=\min_{t\in[T+1]}\min_{i<j}\|s_{p,t}^i-s_{p,t}^j\|_2.
\end{equation*}
We plot baseline-minus-ours gaps in Fig.~\ref{fig:dense_task_metric_noise} so that positive values indicate favorable relative performance for our methods.\footnote{In navigation tasks, we use ours-minus-baseline for the gap in minimum distance to preserve the interpretation of positive values.}

\subsection{Dense Estimation Simulations}
We first report the results for the dense estimation simulations. We compare our proposed joint control-estimation design with the baseline using the metrics in Sec.~\ref{sec:eval_metrics}.

These simulations evaluate control stationarity, robot cost, and task-specific performance under minimum-variance estimation.

\begin{figure*}[t]
\centering
\includegraphics[width=\linewidth]{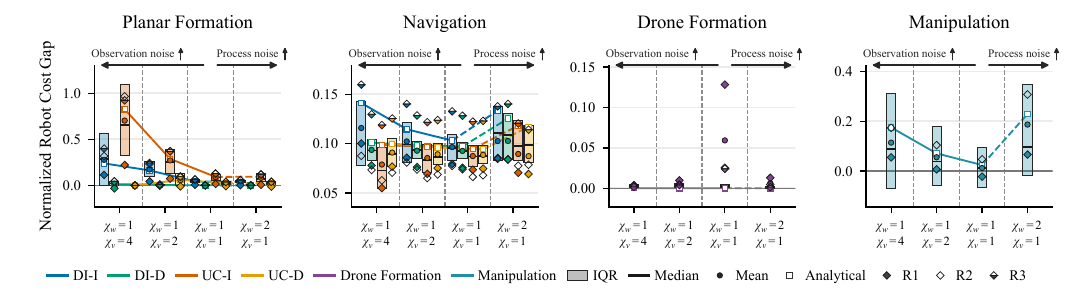}
\caption{
Analytical and realized robot-cost gaps, normalized over horizon $T+1$, across the tested noise levels (baseline minus our proposed method; positive values favor ours).
Diamond markers with different fills show the mean performance gap for each robot. 
The square markers show the analytical cost gap averaged over robots.
The circle markers show the realized cost gap averaged over robots and 100 Monte Carlo rollouts.
Boxes show the interquartile range (IQR) of the rollout cost gaps.
}
\label{fig:dense_nominal_noise}
\end{figure*}

\begin{figure*}[t]
\centering
\includegraphics[width=\linewidth]{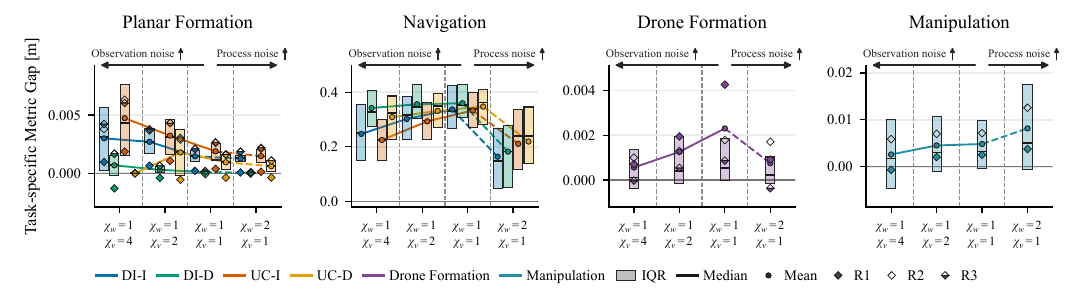}
\caption{
Realized task-specific metric gap between our method and the baseline at different noise levels (positive values favor our proposed method).
Marker styles match Fig.~\ref{fig:dense_nominal_noise}.
}
\label{fig:dense_task_metric_noise}
\end{figure*}

\noindent\textbf{Baseline.}
The baseline combines deterministic LQ (iLQ) feedback Nash policies with stepwise minimum-variance estimators computed for that fixed policy, as in~\cite{qiu2026linear}. 

\noindent\textbf{Runtime analysis.} We ran the offline solver on an AMD Ryzen 9 7950X workstation (16 cores, 32 hardware threads, approximately 125 GiB RAM).
For planar formation tasks, each iteration of Algorithm~\ref{alg:nested_moment} takes \SI{0.95}{\second}. 
For navigation tasks, each iteration of Algorithm~\ref{alg:nested_moment} takes \SI{0.30}{\second}. 
For drone formation tasks, each iteration of Algorithm~\ref{alg:nested_moment} takes \SI{7.30}{\second}. 
For manipulation tasks, each iteration takes \SI{1.13}{\second}.
For planar formation and navigation tasks, optimization time is  proportional to the game horizon $T$ ($188$ vs. $60$).
Nonlinear dynamics slow down optimization greatly in drone formation due to the additional outer iLQ iterations required in Algorithm \ref{alg:ilq_alt_fne}.
Such slowdown is less severe in manipulation tasks than drone formation tasks as the MuJoCo \cite{todorov2012mujoco} simulator provides a built-in linearized dynamics for robot arms, which speeds up optimization.

\noindent\textbf{How closely do the computed policies satisfy the stationarity equations?}
Figure~\ref{fig:dense_stationarity_noise} shows that the joint design generally reduces the coupled-control stationarity residual compared to the baseline. This reveals that the deterministic-policy baseline does not generally satisfy the coupled-control stationarity equations evaluated here, while the proposed design yields orders-of-magnitude smaller final residuals \eqref{eqn:dense_stationarity_residual} in the tested cases.

\noindent\textbf{How does performance vary across noise scales?}
For each task, we scale the process and observation noise standard deviations by factors $\chi_w$ and $\chi_v$, respectively. We evaluate both methods' performance under ${(\chi_w,\chi_v)\in\{(1,1),(1,2),(1,4),(2,1)\}}$. At each noise level, the policy and estimators are re-optimized since the covariance varies.
Our method outperforms the baseline in all tasks with all models at every noise level. 
In most tasks, we observe an increasing mean performance gap as the process or observation noise increases, indicating that the mean improvement is larger in high-noise scenarios. 

\subsection{Sparse Estimator Simulations}
\label{sec:experiment_sparse}

We next investigate how reducing the number of sensor observations affects robot cost and whether policy-adaptive regularization improves this trade-off.

\begin{figure*}[t]
\centering
\includegraphics[width=\linewidth]{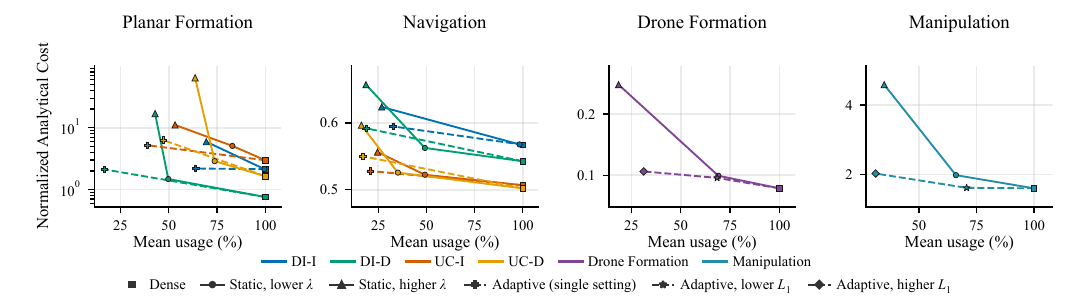}
\caption{Analytical robot cost versus mean sensor-block usage for dense estimation, two static regularization levels, and policy-adaptive regularization.
Sensor-block usage is the fraction of activated sensor-blocks over all sensor-blocks over the horizon (cf. Fig.~\ref{fig:sensor_usage}).
The drone formation and manipulation panels each show two adaptive settings. Solid lines connect the dense design to static designs, while dashed lines connect the dense design to adaptive designs; lower and farther left indicate a better cost-usage trade-off.}
\label{fig:sparse_pareto}
\end{figure*}

\begin{figure*}[t]
\centering
\includegraphics[width=\linewidth]{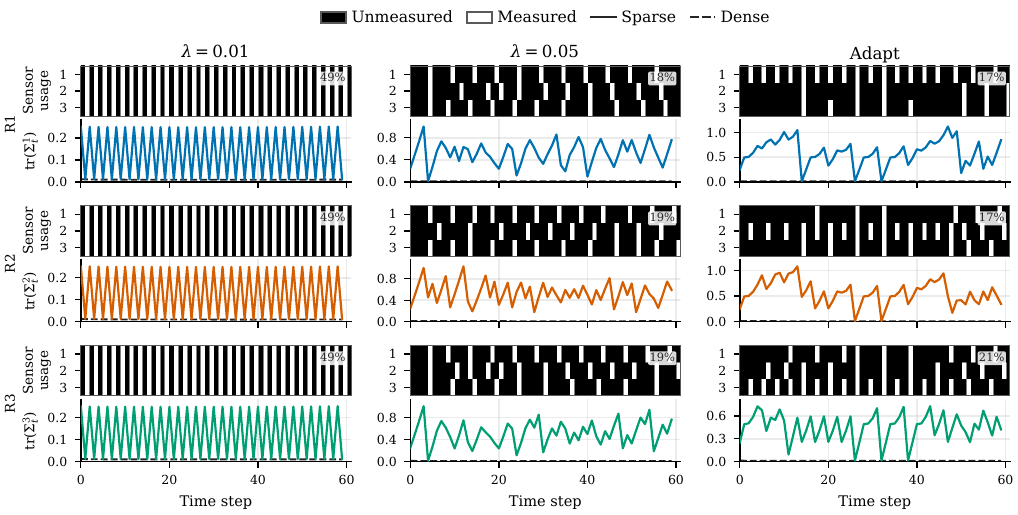}
\caption{Sensor-block selection and estimation-error covariance in the decentralized double-integrator navigation model (DI-D). Columns compare static $\lambda=0.01$, static $\lambda=0.05$, and policy-adaptive regularization; rows correspond to the three robots. In each row, the upper strip marks observed blocks in white and omitted blocks in black over time, with mean block usage shown at the right. The lower plot shows the trace of that robot's estimation-error covariance for the sparse estimator (solid) and dense estimator (dashed). Stronger sparsification lowers usage while increasing estimation uncertainty.}
\label{fig:sensor_usage}
\end{figure*}

\noindent\textbf{Setup and baselines.}
For each task, we compute the dense policy at nominal noise, $(\chi_w,\chi_v)=(1,1)$, and hold its feedback gains $(\Gamma,\alpha)$ fixed throughout the sparse comparison.
We compare the policy-adaptive group-lasso estimator in~\eqref{eqn:adaptive_lambda} with static group-lasso regularization at two fixed values of $\lambda$ and with dense minimum-variance estimation.
All sparse estimator designs apply group-lasso regularization followed by corrective reset with $r_{\text{th}}=0.2$.
Figure~\ref{fig:sparse_pareto} reports average analytical costs over robots; Fig.~\ref{fig:sensor_usage} shows the retained blocks and estimation-error covariance traces over time for the navigation task with the DI-D model.

\noindent\textbf{How many observation blocks does the sparse estimator retain?}
Figure~\ref{fig:sparse_pareto} shows that sparse estimation can substantially reduce observation use while preserving much of the dense design's performance.
For DI-I planar formation, adaptive regularization retains $64.0\%$ of observation blocks while increasing analytical cost from $2.129$ to $2.185$.
More aggressive sparsification generally incurs a larger cost. For example, in drone formation, increasing the static weight from $0.0001$ to $0.001$ reduces usage from $69.0\%$ to $18.5\%$, while cost rises from $0.099$ to $0.248$.

\noindent\textbf{Does policy adaptation improve the cost--usage trade-off?}
At comparable sensor usage, policy-adaptive regularization can outperform static regularization.
In DI-D navigation, both adaptive regularization and static $\lambda=0.05$ retain $18.6\%$ of observation blocks, but their analytical costs are $0.592$ and $0.657$, respectively.
In manipulation, adaptive $L_1=0.02$ reduces usage from $34.9\%$ to $31.2\%$ and cost from $4.585$ to $2.032$ relative to static $\lambda=0.005$.
The trade-off remains sensitive to the regularization setting: for UC-I planar formation, adaptation reduces usage from $82.9\%$ to $38.9\%$ relative to static $\lambda=0.01$, with a small cost increase from $5.068$ to $5.206$.

\noindent\textbf{How does adaptation change observation selection?}
Figure~\ref{fig:sensor_usage} examines DI-D navigation in more detail.
Static $\lambda=0.01$ retains about $49\%$ of observation blocks for each robot; increasing $\lambda$ to $0.05$ reduces usage to $18$--$19\%$ and produces larger covariance peaks.
The adaptive estimator assigns different observation schedules to the robots, retaining $17.5\%$, $17.5\%$, and $20.8\%$ for robots~1--3, respectively.
Its covariance traces are not uniformly lower than those of the strongly regularized static estimator, yet its robot cost is lower at the same mean usage.
This comparison illustrates the value of selecting observations according to their influence on feedback decisions.

\subsection{Robot Hardware Demonstration}
\begin{figure*}[t]
\centering
\includegraphics[width=\linewidth]{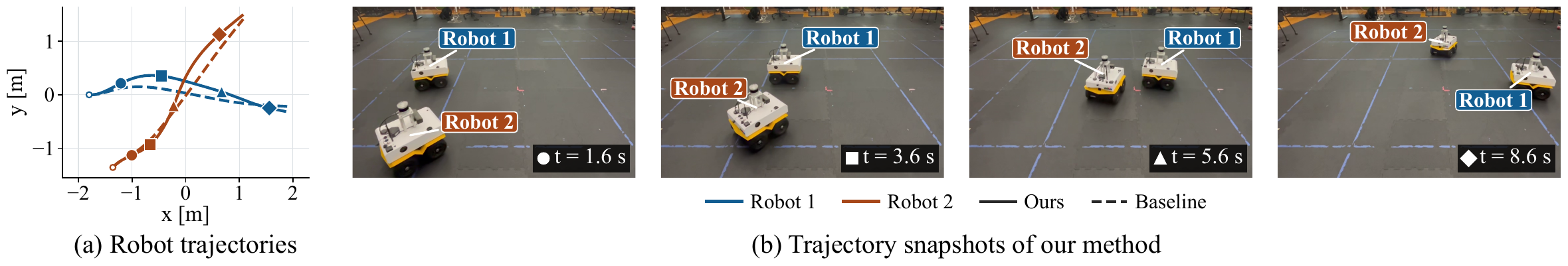}
\caption{Navigation with two Clearpath Jackals. (a) Plotted robot trajectories (solid: ours, dashed: baseline). (b) Time-stamped snapshots of the robots executing the proposed controller and estimator.}
\label{fig:hardware}
\end{figure*}

We implemented our proposed method as a Robot Operating System (ROS) package and deployed it on two Clearpath Jackals (UC dynamics) for the navigation task described in Sec.~\ref{sec:experiment_navigation}, with $\Delta t=\SI{0.2}{\second}$ and $T=50$ (Fig.~\ref{fig:hardware}).
Offline optimization averaged 1.67 s per iteration of Algorithm \ref{alg:nested_moment} on a Dell Latitude 5420 with an Intel Core i7-1185G7 and 15 GiB of RAM.
A Vicon system supplies the joint state used to construct an independently noise-corrupted I-Obs stream for each robot. 
Optimization for both robots run on the laptop; each uses its own observation stream and private estimate. The laptop sends the resulting commands to the two Jackals at $\SI{5}{\hertz}$.

Our method reduces each robot's cost from $1.239$ to $0.409$ (normalized over the time horizon) and increases minimum separation from \SI{0.290}{\metre} to \SI{0.894}{\metre}.
This validates our method's advantage over the baseline in both reducing robot cost and improving navigation performance.

\section{Conclusion}
\label{sec:conclusion}
Motivated by the challenge of multi-robot coordination in communication-restricted environments, we proposed an iterative control and estimation optimization framework to find affine feedback Nash policies and corresponding minimum-variance estimators for linear-quadratic stochastic dynamic games with individual noisy sensor observations.
We extended this framework to nonlinear setups using iLQ approximation.
To reduce sensor usage, we applied the policy-adaptive group-lasso estimator of Qiu et al. \cite{qiu2026linear} that provides a tunable trade-off between robot cost and observation use.
Across four multi-robot simulation tasks, the proposed method reduces stationarity residuals and robot costs relative to deterministic Nash baselines.
Future work will investigate more challenging scenarios such as stress test in larger-scale robot systems with more aggressively nonlinear dynamics and measurements.

\appendices
\section{Proofs of the Main Results}
\label{app:proofs}

\subsection{Proof of Proposition~\ref{prop:quadratic_cost_to_go}}
\label{app:proof_quadratic_cost_to_go}
\begin{proof}
We prove this by backward induction on $\tau$.
For $\tau=T+1$:

\begin{equation*}
    \begin{aligned}
        \mathcal{J}_{T+1}^i(\textbf{z}_{T+1})&=\mathbb{E}\left.\left[\frac{1}{2}\textbf{x}_{T+1}^\top Q_{T+1}^i\textbf{x}_{T+1}+q_{T+1}^{i\top}\textbf{x}_{T+1}+\ell_{\mathrm{const},T+1}^i\right|\textbf{z}_{T+1}\right]\\
        &=\frac{1}{2}\textbf{z}_{T+1}^\top E_0^\top Q_{T+1}^iE_0\textbf{z}_{T+1}+q_{T+1}^{i\top}E_0\textbf{z}_{T+1}+\ell_{\mathrm{const},T+1}^i,
    \end{aligned}
\end{equation*}
i.e., $P_{T+1}^i=E_0^\top Q_{T+1}^iE_0$, $p_{T+1}^i=E_0^\top q_{T+1}^i$, and $c_{T+1}^i=2\ell_{\mathrm{const},T+1}^i$.
Suppose $\mathcal{J}_\tau^i(\textbf{z}_\tau)$ is quadratic for $t+1\leq\tau\leq T+1$. For $\tau=t$, it follows from $\textbf{x}_t=E_0\textbf{z}_t$ and $\textbf{u}_t=-\Gamma_{\textbf{z},t}\textbf{z}_t-\alpha_t$ that
    
\begin{equation*}
    \begin{aligned}
        &\mathcal{J}_t^i(\textbf{z}_t)=\mathbb{E}\left.\left[\ell_t^i(\textbf{x}_t,\textbf{u}_t)+\mathcal{J}_{t+1}^i(\textbf{z}_{t+1})\right|\textbf{z}_t\right]\\
        =\ &\frac{1}{2}\textbf{z}_t^\top E_0^\top Q_t^iE_0\textbf{z}_t+q_t^{i\top}E_0\textbf{z}_t+\frac{1}{2}(\Gamma_{\textbf{z},t}\textbf{z}_t+\alpha_t)^\top R_t^i(\Gamma_{\textbf{z},t}\textbf{z}_t+\alpha_t)\\
        -\ &r_t^{i\top}(\Gamma_{\textbf{z},t}\textbf{z}_t+\alpha_t)+\ell_{\mathrm{const},t}^i+\mathbb{E}\left.\left[\mathcal{J}_{t+1}^i(\textbf{z}_{t+1})\right|\textbf{z}_t\right].
    \end{aligned}
\end{equation*}
Substituting \eqref{eqn:z_update}, we obtain
    
\begin{equation*}
    \begin{aligned}
        \mathbb{E}\left.\left[\mathcal{J}_{t+1}^i(\textbf{z}_{t+1})\right|\textbf{z}_t\right]&=\frac{1}{2}\textbf{z}_t^\top\mathcal{A}_{\textbf{z},t}^\top P_{t+1}^i\mathcal{A}_{\textbf{z},t}\textbf{z}_t\\
        &+\left[(f_{\text{const},t}-B_t\alpha_t)^\top E_0 P_{t+1}^i+p_{t+1}^{i\top}\right]\mathcal{A}_{\textbf{z},t}\textbf{z}_t\\
        &+\frac{1}{2}c_{t+1}^i+\frac{1}{2}\text{tr}(\mathcal{K}_{\omega,t+1}^\top P_{t+1}^i\mathcal{K}_{\omega,t+1}\Omega_t)\\
        &+\frac{1}{2}(f_{\text{const},t}-B_t\alpha_t)^\top E_0P_{t+1}^iE_0^\top(f_{\text{const},t}-B_t\alpha_t)\\
        &+p_{t+1}^{i\top}E_0^\top(f_{\text{const},t}-B_t\alpha_t).
    \end{aligned}
\end{equation*}
This expression is quadratic with respect to $\textbf{z}_t$.
Collecting coefficients of $\textbf{z}_t$ yields the backward recursion~\eqref{eqn:Ppc_recursion}, which completes the proof.
\end{proof}

\subsection{Proof of Theorem~\ref{thm:coupled_expected_stationarity}}
\label{app:proof_coupled_expected_stationarity}
\begin{proof}
Write $\langle X,Y\rangle\coloneqq\operatorname{tr}(X^\top Y)$ and recall

\begin{equation*}
\begin{aligned}
    \Gamma_{\textbf{z},t}&=\sum_{k=1}^N \Pi^k\Gamma_t D_{0k},\\
\mathcal{A}_{\textbf{z},t}&=\mathcal{F}_{0,t}+\sum_{j=1}^N L_{j0,t}\Gamma_t^j D_{0j}+\sum_{j=1}^N\sum_{\ell\neq j}L_{j\ell,t}\Gamma_t^j D_{j\ell}.
\end{aligned}
\end{equation*}
With $(P_{t+1}^i,p_{t+1}^i,\mu_t,\Xi_t)$ held fixed in the backward stage solve, the part of $\bar{\mathcal{J}}_t^i$ in \eqref{eqn:expected_augmented_cost} that depends on $(\Gamma_t,\alpha_t)$ is obtained by substituting the recursions in Proposition \ref{prop:quadratic_cost_to_go} for $(P_t^i,p_t^i,c_t^i)$:

\begin{equation*}
\begin{aligned}
\bar{\mathcal{J}}_t^i
&=\tfrac12\operatorname{tr}(\Gamma_{\textbf{z},t}^\top R_t^i \Gamma_{\textbf{z},t}\Xi_t)
+\mu_t^\top \Gamma_{\textbf{z},t}^\top(R_t^i\alpha_t-r_t^i)
+\tfrac12\alpha_t^\top R_t^i\alpha_t-r_t^{i\top}\alpha_t\\
&+\tfrac12\operatorname{tr}(\mathcal{A}_{\textbf{z},t}^\top P_{t+1}^i\mathcal{A}_{\textbf{z},t}\Xi_t)
\\
&+\left(P_{t+1}^iE_0^\top(f_{\mathrm{const},t}-B_t\alpha_t)+p_{t+1}^i\right)^\top\mathcal{A}_{\textbf{z},t}\mu_t\\
&+\tfrac12(f_{\mathrm{const},t}-B_t\alpha_t)^\top E_0P_{t+1}^iE_0^\top(f_{\mathrm{const},t}-B_t\alpha_t)\\
&+p_{t+1}^{i\top}E_0^\top(f_{\mathrm{const},t}-B_t\alpha_t)
+\text{$(\Gamma_t,\alpha_t)$-independent terms}.
\end{aligned}
\end{equation*}

\noindent\textit{\textbf{Variation of $\bar{\mathcal{J}}_t^i$ with respect to $\Gamma_t^i$.}}
A variation $\delta\Gamma_t^i$ induces

\begin{equation*}
\delta \Gamma_{\textbf{z},t}=\Pi^i(\delta\Gamma_t^i)D_{0i},\quad
\delta\mathcal{A}_{\textbf{z},t}=L_{i0,t}(\delta\Gamma_t^i)D_{0i}+\sum_{\ell\neq i}L_{i\ell,t}(\delta\Gamma_t^i)D_{i\ell}.
\end{equation*}

\noindent\romannumeral1) The stage-control quadratic depends on $\Gamma_t$ only through $\Gamma_{\textbf{z},t}$.
Expanding

\begin{equation*}
\Gamma_{\textbf{z},t}^\top R_t^i \Gamma_{\textbf{z},t}
=\sum_{j=1}^N\sum_{k=1}^N D_{0j}^\top\Gamma_t^{j\top}R_{jk,t}^i\Gamma_t^k D_{0k}
\end{equation*}
and taking the first variation of $\tfrac12\operatorname{tr}(\Gamma_{\textbf{z},t}^\top R_t^i \Gamma_{\textbf{z},t}\Xi_t)$ with respect to $\Gamma_t^i$ gives

\begin{equation*}
\left\langle
\sum_{k=1}^N R_{ik,t}^i\Gamma_t^k D_{0k}\Xi_t D_{0i}^\top,
\delta\Gamma_t^i
\right\rangle.
\end{equation*}

\noindent\romannumeral2) The closed-loop quadratic $\tfrac12\operatorname{tr}(\mathcal{A}_{\textbf{z},t}^\top P_{t+1}^i\mathcal{A}_{\textbf{z},t}\Xi_t)$ varies as

\begin{equation*}
\left\langle
L_{i0,t}^\top P_{t+1}^i\mathcal{A}_{\textbf{z},t}\Xi_t D_{0i}^\top
+\sum_{\ell\neq i}L_{i\ell,t}^\top P_{t+1}^i\mathcal{A}_{\textbf{z},t}\Xi_t D_{i\ell}^\top,
\delta\Gamma_t^i
\right\rangle.
\end{equation*}

\noindent\romannumeral3) The mean-dependent cross term $\mu_t^\top \Gamma_{\textbf{z},t}^\top(R_t^i\alpha_t-r_t^i)$ contributes

\begin{equation*}
\left\langle
\left(\sum_{k=1}^N R_{ik,t}^i\alpha_t^k-r_{i,t}^i\right)\mu_t^\top D_{0i}^\top,
\delta\Gamma_t^i
\right\rangle,
\end{equation*}
while $\left(P_{t+1}^iE_0^\top(f_{\mathrm{const},t}-B_t\alpha_t)+p_{t+1}^i\right)^\top\mathcal{A}_{\textbf{z},t}\mu_t$ contributes

\begin{equation*}
\begin{aligned}
    &\left\langle
L_{i0,t}^\top\left(P_{t+1}^iE_0^\top(f_{\mathrm{const},t}-B_t\alpha_t)+p_{t+1}^i\right)\mu_t^\top D_{0i}^\top,
\delta\Gamma_t^i
\right\rangle
\\+&\sum_{\ell\neq i}
\left\langle
L_{i\ell,t}^\top\left(P_{t+1}^iE_0^\top(f_{\mathrm{const},t}-B_t\alpha_t)+p_{t+1}^i\right)\mu_t^\top D_{i\ell}^\top,
\delta\Gamma_t^i
\right\rangle.
\end{aligned}
\end{equation*}
Collecting \romannumeral1)-\romannumeral3), setting the coefficient of every $\delta\Gamma_t^i$ to zero, and vectorizing with $\operatorname{vec}(AXB)=(B^\top\otimes A)\operatorname{vec}(X)$ and $\operatorname{vec}(A\alpha v^\top)=(v\otimes A)\alpha$ yields the block row in $\Gamma_t^i$

\begin{equation*}
\sum_{k}H_{ik,t}\operatorname{vec}(\Gamma_t^k)-\sum_{k}\mathfrak{Y}_{ik,t}\alpha_t^k=b_{i,t},
\end{equation*}
with $H_{ik,t}$, $\mathfrak{Y}_{ik,t}$, and $b_{i,t}$ as in \eqref{eqn:coupled_blocks}.\\

\noindent\textit{\textbf{Variation of $\bar{\mathcal{J}}_t^i$ with respect to $\alpha_t^i$.}} Proceeding as for $\Gamma_t^i$, the first variation in $\delta\alpha_t^i$ is

\begin{equation*}
\left\langle
\sum_{k}\mathfrak{Z}_{ik,t}\alpha_t^k-\sum_{k}G_{ik,t}\operatorname{vec}(\Gamma_t^k)-\beta_{i,t},
\delta\alpha_t^i
\right\rangle.
\end{equation*}
Setting $\nabla_{\alpha_t^i}\bar{\mathcal{J}}_t^i=\textbf{0}$ yields the block row in $\alpha_t^i$

\begin{equation*}
\sum_{k}\mathfrak{Z}_{ik,t}\alpha_t^k-\sum_{k}G_{ik,t}\operatorname{vec}(\Gamma_t^k)=\beta_{i,t},
\end{equation*}
with $\mathfrak{Z}_{ik,t}$, $G_{ik,t}$, and $\beta_{i,t}$ as in \eqref{eqn:coupled_blocks}.\\
Collecting the coefficients of the two block rows gives the explicit expressions

\begin{subequations}
\begin{align}
H_{ik,t}&=
(D_{0i}\Xi_t D_{0k}^\top)\otimes R_{ik,t}^i
+(D_{0i}\Xi_t D_{0k}^\top)\otimes(L_{i0,t}^\top P_{t+1}^i L_{k0,t})\notag\\
&+\sum_{\ell\neq i}(D_{0i}\Xi_t D_{k\ell}^\top)\otimes(L_{i0,t}^\top P_{t+1}^i L_{k\ell,t})\notag\\
&+\sum_{\ell\neq i}(D_{i\ell}\Xi_t D_{0k}^\top)\otimes(L_{i\ell,t}^\top P_{t+1}^i L_{k0,t})\notag\\
&+\sum_{\ell\neq i}\sum_{r\neq k}(D_{i\ell}\Xi_t D_{kr}^\top)\otimes(L_{i\ell,t}^\top P_{t+1}^i L_{kr,t}),\\
b_{i,t}&=
-\operatorname{vec}\left(L_{i0,t}^\top P_{t+1}^i\mathcal{F}_{0,t}\Xi_t D_{0i}^\top\right)
-\sum_{\ell\neq i}\operatorname{vec}\left(L_{i\ell,t}^\top P_{t+1}^i\mathcal{F}_{0,t}\Xi_t D_{i\ell}^\top\right)\notag\\
&-\operatorname{vec}\left(
r_{i,t}^i\mu_t^\top D_{0i}^\top
+L_{i0,t}^\top\left(P_{t+1}^iE_0^\top f_{\mathrm{const},t}+p_{t+1}^i\right)\mu_t^\top D_{0i}^\top\right.\notag\\
&\left.+\sum_{\ell\neq i}L_{i\ell,t}^\top\left(P_{t+1}^iE_0^\top f_{\mathrm{const},t}+p_{t+1}^i\right)\mu_t^\top D_{i\ell}^\top\right),\\
\mathfrak{Y}_{ik,t}&=
(D_{0i}\mu_t)\otimes R_{ik,t}^i
+(D_{0i}\mu_t)\otimes(L_{i0,t}^\top P_{t+1}^iE_0^\top B_t^k)\notag\\
&+\sum_{\ell\neq i}(D_{i\ell}\mu_t)\otimes(L_{i\ell,t}^\top P_{t+1}^iE_0^\top B_t^k),\\
\mathfrak{Z}_{ik,t}&=
R_{ik,t}^i+B_t^{i\top} E_0P_{t+1}^iE_0^\top B_t^k,\\
\beta_{i,t}&=r_{i,t}^i+
B_t^{i\top} E_0
\left[P_{t+1}^i(\mathcal{F}_{0,t}\mu_t+E_0^\top f_{\mathrm{const},t})
+p_{t+1}^i\right],\\
G_{ik,t}&=
(D_{0k}\mu_t)^\top\otimes R_{ik,t}^i
+(D_{0k}\mu_t)^\top\otimes\left(B_t^{i\top} E_0P_{t+1}^i L_{k0,t}\right)\notag\\
&+\sum_{r\neq k}(D_{kr}\mu_t)^\top\otimes\left(B_t^{i\top} E_0P_{t+1}^i L_{kr,t}\right).
\end{align}
\label{eqn:coupled_blocks}
\end{subequations}

\noindent\textbf{\textit{Coupled linear systems for $\Gamma_t^i$ and $\alpha_t^i$}.}
Stacking the first-order conditions in $(\Gamma_t^i,\alpha_t^i)$ over $i\in[N]$ yields \eqref{eqn:coupled_gain_alpha_system}, which completes the proof.
\end{proof}

\subsection{Proof of Corollary~\ref{cor:local_nash_affine}}
\label{app:proof_local_nash_affine}
\begin{proof}
Equation~\eqref{eqn:coupled_gain_alpha_system} is the stacked first-order condition $\nabla_{(\Gamma_t^i,\alpha_t^i)}\bar{\mathcal{J}}_t^i=\textbf{0}$ for every $i$.
The moments that enter that linear system are $(\mu_t^\star,\Xi_t^\star)$, the statistics of $\textbf{z}_t$ generated by the candidate itself through~\eqref{eqn:moment_update}; they are held fixed while differentiating $\bar{\mathcal{J}}_t^i$ in $(\Gamma_t^i,\alpha_t^i)$.
Positive definiteness of the Hessian with respect to $(\Gamma_t^i,\alpha_t^i)$ is the second-order sufficient condition for a strict local minimizer of $\bar{\mathcal{J}}_t^i$ in $(\Gamma_t^i,\alpha_t^i)$.
Hence no agent has a profitable local unilateral deviation in the affine class~\eqref{eqn:actual_single_ctrl}.
\end{proof}
\section{Special Cases for Stochastic Dynamic Games}
\label{app:special_case}
\subsection{Single-agent LQG reduction}
\label{app:proof_single_agent_lqg}
\begin{corollary}[Single-agent LQG reduction]
\label{cor:single_agent_lqg}
For $N=1$ there is no control-imputation mismatch, so $\hat{\textbf{x}}_t$ is the Kalman conditional-mean estimate ($\mathbb{E}[\textbf{e}_t\mid\hat{\textbf{x}}_t]=\textbf{0}$).
Write

\begin{equation}
P_{\textbf{xx},t+1}\coloneqq E_0P_{t+1}E_0^\top,\qquad
p_{\textbf{x},t+1}\coloneqq E_0p_{t+1},
\label{eqn:lqg_Pxx}
\end{equation}
and assume $R_t+B_t^\top P_{\textbf{xx},t+1}B_t$ is nonsingular.
Then \eqref{eqn:coupled_gain_alpha_system} reduces to the classical LQG (certainty-equivalent) affine law $u_t=-\Gamma_t\hat{\textbf{x}}_t-\alpha_t$ with

\begin{equation}
\begin{aligned}
\Gamma_t&=(R_t+B_t^\top P_{\textbf{xx},t+1}B_t)^{-1}B_t^\top P_{\textbf{xx},t+1}A_t,\\
\alpha_t&=(R_t+B_t^\top P_{\textbf{xx},t+1}B_t)^{-1}
\left[r_t+B_t^\top(P_{\textbf{xx},t+1}f_{\mathrm{const},t}+p_{\textbf{x},t+1})\right].
\end{aligned}
\label{eqn:lqg_gamma_alpha}
\end{equation}
\end{corollary}

\begin{proof}
For $N=1$ the only active channel is the estimate channel, with $L_{10,t}=-E_0^\top B_t$ and $D_{01}\textbf{z}_t=\hat{\textbf{x}}_t$.
Moreover, $\mathcal{F}_{0,t}=E_0^\top A_tE_0$ on the state block that enters the cost, and $\Gamma_{\textbf{z},t}\textbf{z}_t=\Gamma_t\hat{\textbf{x}}_t$ under $u_t=-\Gamma_{\textbf{z},t}\textbf{z}_t-\alpha_t$.
The coefficients in~\eqref{eqn:lqg_Pxx} extract that state block of $(P_{t+1},p_{t+1})$.\\

\noindent\textit{\textbf{Variation of $\bar{\mathcal{J}}_t$ with respect to $\alpha_t$.}}
From \eqref{eqn:coupled_blocks}, after extracting the state block,

\begin{equation*}
\mathfrak{Z}_t=R_t+B_t^\top P_{\textbf{xx},t+1}B_t,\qquad
\beta_t=r_t+B_t^\top\left(P_{\textbf{xx},t+1}f_{\mathrm{const},t}+p_{\textbf{x},t+1}\right),
\end{equation*}
so $\mathfrak{Z}_t\alpha_t=\beta_t$ is exactly the second line of \eqref{eqn:lqg_gamma_alpha}.\\

\noindent\textit{\textbf{Variation of $\bar{\mathcal{J}}_t$ with respect to $\Gamma_t$.}}
Since $\hat{\textbf{x}}_t$ is the conditional mean, the error $\textbf{e}_t=\textbf{x}_t-\hat{\textbf{x}}_t$ is uncorrelated with $\hat{\textbf{x}}_t$, so $\operatorname{Cov}(\textbf{x}_t,\hat{\textbf{x}}_t)=\operatorname{Cov}(\hat{\textbf{x}}_t)$ and

\begin{equation*}
D_{01}\Xi_t D_{01}^\top=\mathbb{E}[\hat{\textbf{x}}_t\hat{\textbf{x}}_t^\top].
\end{equation*}
The variation in \romannumeral1)-\romannumeral2) of the proof of Theorem \ref{thm:coupled_expected_stationarity} therefore reduces to

\begin{equation*}
\left\langle
\left(R_t+B_t^\top P_{\textbf{xx},t+1}B_t\right)\Gamma_t\operatorname{Cov}(\hat{\textbf{x}}_t)-B_t^\top P_{\textbf{xx},t+1}A_t\operatorname{Cov}(\hat{\textbf{x}}_t),
\delta\Gamma_t
\right\rangle
\end{equation*}
plus mean-direction residuals involving $\alpha_t$.
Substituting the solution for $\alpha_t$ cancels those residuals and leaves

\begin{equation*}
\left(R_t+B_t^\top P_{\textbf{xx},t+1}B_t\right)\Gamma_t\operatorname{Cov}(\hat{\textbf{x}}_t)=B_t^\top P_{\textbf{xx},t+1}A_t\operatorname{Cov}(\hat{\textbf{x}}_t).
\end{equation*}
Right cancellation of $\operatorname{Cov}(\hat{\textbf{x}}_t)$ yields the first line of \eqref{eqn:lqg_gamma_alpha}.
\end{proof}

\subsection{Deterministic LQ game reduction}
\label{app:proof_deterministic_nash}
\begin{corollary}[Deterministic LQ game reduction]
\label{cor:deterministic_nash}
Suppose the initial state is deterministic and there is no process or measurement noise ($\Sigma_{\textbf{x},1}^-=\textbf{0}$, $W_t=\textbf{0}$, $V_t^i=\textbf{0}$), so that $\textbf{e}_t^i\equiv\textbf{0}$ for all $i\in[N]$.
Write

\begin{equation*}
P_{\textbf{xx},t+1}^i\coloneqq E_0P_{t+1}^iE_0^\top,\qquad
p_{\textbf{x},t+1}^i\coloneqq E_0p_{t+1}^i.
\label{eqn:det_Pxx}
\end{equation*}
Then, writing

\begin{equation}
\Xi_t=\mu_t\mu_t^\top,
\label{eqn:det_Xi}
\end{equation}
the $\Gamma$ block of \eqref{eqn:coupled_gain_alpha_system} is determined along the mean trajectory, and \eqref{eqn:coupled_gain_alpha_system} reduces to the classical deterministic feedback Nash equations in both $\Gamma_t$ and $\alpha_t$ for each $i\in[N]$:

\begin{equation}
\begin{aligned}
\sum_{k=1}^N
\left(R_{ik,t}^i + B_t^{i\top} P_{\textbf{xx},t+1}^i B_t^k\right)\Gamma_t^k
&=B_t^{i\top} P_{\textbf{xx},t+1}^i A_t,\\
\sum_{k=1}^N
\left(R_{ik,t}^i + B_t^{i\top} P_{\textbf{xx},t+1}^i B_t^k\right)\alpha_t^k
&=r_{i,t}^i
+B_t^{i\top}(P_{\textbf{xx},t+1}^i f_{\mathrm{const},t}+p_{\textbf{x},t+1}^i).
\end{aligned}
\label{eqn:deterministic_lq_gain}
\end{equation}
In particular, $u_t^i=-\Gamma_t^i\textbf{x}_t-\alpha_t^i$, the standard full-state deterministic Nash strategy. See Appendix~\ref{app:proof_deterministic_nash} for a detailed proof.
\end{corollary}

\begin{proof}
With $W_t=\textbf{0}$, $V_t^i=\textbf{0}$, and $\textbf{e}_t\equiv\textbf{0}$, one has $\textbf{z}_t=E_0^\top\textbf{x}_t$.
Every error channel disappears, and the only surviving channels are $(L_{j0,t},D_{0j})$ with

\begin{equation*}
L_{j0,t}=-E_0^\top B_t^j,\qquad
D_{0j}E_0^\top=I,\qquad
\mathcal{F}_{0,t}=E_0^\top A_tE_0,
\label{eqn:det_channels}
\end{equation*}
and the joint map reduces to the full-state law $\textbf{u}_t=-\Gamma_t\textbf{x}_t-\alpha_t$.
The closed-loop state evolves deterministically, so~\eqref{eqn:det_Xi} holds.\\

\noindent\textit{\textbf{Variation of $\bar{\mathcal{J}}_t^i$ with respect to $\alpha_t^i$.}}
Specializing the variations \romannumeral1)-\romannumeral3) in the proof of Theorem \ref{thm:coupled_expected_stationarity}--or directly the blocks in \eqref{eqn:coupled_blocks}--yields

\begin{equation*}
\sum_{k}\left(R_{ik,t}^i+B_t^{i\top} P_{\textbf{xx},t+1}^i B_t^k\right)\alpha_t^k
=r_{i,t}^i+B_t^{i\top}(P_{\textbf{xx},t+1}^i f_{\mathrm{const},t}+p_{\textbf{x},t+1}^i),
\end{equation*}
i.e., the second block of \eqref{eqn:deterministic_lq_gain}.\\

\noindent\textit{\textbf{Variation of $\bar{\mathcal{J}}_t^i$ with respect to $\Gamma_t^i$.}}
Likewise, the control-cost and closed-loop variations reduce to

\begin{equation*}
H_{ik,t}=(D_{0i}\Xi_t D_{0k}^\top)\otimes R_{ik,t}^i
+(D_{0i}\Xi_t D_{0k}^\top)\otimes\left(B_t^{i\top} P_{\textbf{xx},t+1}^i B_t^k\right),
\end{equation*}
since $L_{k0,t}=-E_0^\top B_t^k$ and $D_{0k}E_0^\top=I$.
With~\eqref{eqn:det_Xi}, $D_{0i}\Xi_t D_{0k}^\top=(D_{0i}\mu_t)(D_{0k}\mu_t)^\top$, so each factor is an outer product of the mean state.
Substituting the solution for $\alpha_t$ cancels the mean residuals and leaves

\begin{equation*}
\sum_{k}\left(R_{ik,t}^i+B_t^{i\top} P_{\textbf{xx},t+1}^i B_t^k\right)\Gamma_t^k
=B_t^{i\top} P_{\textbf{xx},t+1}^i A_t,
\end{equation*}
which is the first block of \eqref{eqn:deterministic_lq_gain}.
\end{proof}

These reductions connect the construction to classical single-agent LQG control and deterministic full-state games. Under individual observations, however, the preceding equilibrium interpretation remains restricted to the affine policy class and sufficient conditions (Corollary \ref{cor:local_nash_affine}).
\section{Experimental Parameters}
\label{app:experiment_parameters}

This section collects the parameters used for reproducing robot tasks in Sec.~\ref{sec:experiment}.
Table~\ref{tab:agent_cost_weights} lists weights for robot cost.
Table~\ref{tab:nominal_noise} lists the standard deviations for initial states and noise. 
Table~\ref{tab:distance_noise} lists the values of distance-dependent noise scaling term $\alpha_{\text{noise}}$. 
Table~\ref{tab:sparse-settings} lists the regularization settings for sparse estimator design.

In addition, for navigation tasks, the proximity denominator uses $\delta=0.03$. 
In sparse estimator design, the reset threshold uses $r_{\text{th}}=0.2$.

\begin{table}[H]
\centering
\caption{Robot cost weights for each task. 
}
\label{tab:agent_cost_weights}
\scriptsize
\begin{tabular}{llll}
\toprule
Task & Robot $i$ & $\kappa_t^i$ &$\kappa_{T+1}^i$\\
\midrule
\multirow{2}{*}{Planar formation} & $1$ & $[240,\ 0.025]^\top$ & $288$\\
 & $2,3$ & $[150,\ 0.025]^\top$ & $180$\\
Navigation & $1,2,3$ & $[0.3,\ 0.6,\ 0.1]^\top$ & $[0.36,\ 0.6]^\top$\\
Drone formation & $1,2,3$ & $[20,\ 10,\ 0.05]^\top$ & $[100,\ 50]^\top$\\
Manipulation & $1,2$ & $[42,\ 8,\ 40,\ 0.5]^\top$ & $[252,\ 48,\ 240]^\top$\\
\bottomrule
\end{tabular}
\end{table}

\begin{table}[H]
\centering
\caption{Nominal standard-deviation scales $\sigma_0$ at $(\chi_w,\chi_v)=(1,1)$.}
\label{tab:nominal_noise}
\scriptsize
\begin{tabular}{llll}
\toprule
Model \& Task & Initial state & Process noise & Observation noise\\
\midrule
DI & $[\frac{1}{24},\frac{1}{24},0.2,0.2]$ & $[0,0,0.2,0.2]$ & $[\frac{1}{24},\frac{1}{24},0.2,0.2]$\\
UC & $[\frac{1}{24},\frac{1}{24},\frac{4}{75},\frac{1}{5}]$ & $[0,0,\frac{1}{20},\frac{1}{5}]$ & $[\frac{1}{24},\frac{1}{24},\frac{1}{15},\frac{1}{5}]$\\
Drone formation & $[\frac{I_3}{20},\frac{I_3}{20},\frac{I_3}{5},\frac{I_3}{5}]$ & $[0,0,\frac{3}{400},\frac{3}{800}]$ & $[\frac{I_3}{320},\frac{I_3}{320},\frac{I_3}{128},\frac{I_3}{128}]$\\
Manipulation & $[0.04 I_7,0.10I_7]$ & $[\textbf{0},0.04I_7]$ & $[0.06I_7,0.18I_7]$\\
\bottomrule
\end{tabular}
\end{table}

\begin{table}[H]
\centering
\caption{Distance-dependent noise coefficients $\alpha_\text{noise}$ for D-Obs.}
\label{tab:distance_noise}
\scriptsize
\begin{tabular}{lll}
\toprule
Model & Observation block & $a_{\text{noise}}$\\
\midrule
\multirow{2}{*}{DI} & Self & $[\frac{1}{600},\frac{1}{6000},0,0]$\\
 & Inter-robot & $[\frac{1}{600},\frac{1}{6000},\frac{1}{600},\frac{1}{600}]$\\
\multirow{2}{*}{UC} & Self & $[\frac{1}{600},\frac{1}{6000},0,0]$\\
 & Inter-robot & $[\frac{1}{600},\frac{1}{6000},\frac{1}{6000},\frac{1}{600}]$\\
\bottomrule
\end{tabular}
\end{table}

\begin{table}[H]
    \centering
    \caption{Sparse-estimator regularization settings. 
    }
    \label{tab:sparse-settings}
    \scriptsize
    \begin{tabular}{llll}
    \toprule
    Task & Static $\lambda$ & Adaptive $L_1$ & $L_2$\\
    \midrule
    Planar formation & $0.01,\ 0.05$ & $0.7$ & $2$\\
    Navigation & $0.01,\ 0.05$ & $0.1$ & $2$\\
    Drone formation & $0.0001,\ 0.001$ & $0.002,\ 0.005$ & $2$\\
    Manipulation & $0.001,\ 0.005$ & $0.01,\ 0.02$ & $2$\\
    \bottomrule
    \end{tabular}
\end{table}

\section*{Acknowledgment}
This work was sponsored by  NASA under grants 80NSSC21M0071 and 80NSSC24M0070, by the NSF under grants 2336840 and 2409535, by ARL under cooperative agreement W911NF-25-2-0021, by ARO under grant W911NF-26-1-A268, by ONR under grant N00014-24-1-2797, and by AFOSR under grant FA9550-22-1-0403.

The authors would like to thank Xinjie Liu for helpful discussions in paper writing.

%

\balance
\bibliographystyle{IEEEtran}
\bibliography{IEEEabrv,reference}

\begin{thebibliography}{10}
\providecommand{\url}[1]{#1}
\csname url@samestyle\endcsname
\providecommand{\newblock}{\relax}
\providecommand{\bibinfo}[2]{#2}
\providecommand{\BIBentrySTDinterwordspacing}{\spaceskip=0pt\relax}
\providecommand{\BIBentryALTinterwordstretchfactor}{4}
\providecommand{\BIBentryALTinterwordspacing}{\spaceskip=\fontdimen2\font plus
\BIBentryALTinterwordstretchfactor\fontdimen3\font minus \fontdimen4\font\relax}
\providecommand{\BIBforeignlanguage}[2]{{%
\expandafter\ifx\csname l@#1\endcsname\relax
\typeout{** WARNING: IEEEtran.bst: No hyphenation pattern has been}%
\typeout{** loaded for the language `#1'. Using the pattern for}%
\typeout{** the default language instead.}%
\else
\language=\csname l@#1\endcsname
\fi
#2}}
\providecommand{\BIBdecl}{\relax}
\BIBdecl

\bibitem{saboia2024cadre}
M.~Saboia, F.~Rossi, V.~Nguyen, G.~Lim, D.~Aguilar, and J.-P. de~la Croix, ``Cadre moondb: Distributed database for multi-robot information-sharing and map-merging for lunar exploration,'' in \emph{Int. Workshop on Autonomous Agents and Multi-Agent Systems for Space Applications}, 2024.

\bibitem{qiu2026linear}
T.~Qiu, F.~Fotiadis, X.~Liu, C.~Ellis, J.~Milzman, W.~Suttle, U.~Topcu, and D.~Fridovich-Keil, ``Linear-quadratic gaussian games with distributed sparse estimation,'' \emph{IEEE Control Systems Letters}, 2026.

\bibitem{bertsekas2012dynamic}
D.~Bertsekas, \emph{Dynamic programming and optimal control: Volume I}.\hskip 1em plus 0.5em minus 0.4em\relax Athena scientific, 2012, vol.~4.

\bibitem{lewis2017optimal}
F.~L. Lewis, L.~Xie, and D.~Popa, \emph{Optimal and robust estimation: with an introduction to stochastic control theory}.\hskip 1em plus 0.5em minus 0.4em\relax CRC press, 2017.

\bibitem{witsenhausen1968counterexample}
H.~S. Witsenhausen, ``A counterexample in stochastic optimum control,'' \emph{SIAM Journal on Control}, vol.~6, no.~1, pp. 131--147, 1968.

\bibitem{gupta2014common}
A.~Gupta, A.~Nayyar, C.~Langbort, and T.~Basar, ``Common information based markov perfect equilibria for linear-gaussian games with asymmetric information,'' \emph{SIAM Journal on Control and Optimization}, vol.~52, no.~5, pp. 3228--3260, 2014.

\bibitem{gupta2016dynamic}
A.~Gupta, C.~Langbort, and T.~Ba{\c{s}}ar, ``Dynamic games with asymmetric information and resource constrained players with applications to security of cyberphysical systems,'' \emph{IEEE Transactions on Control of Network Systems}, vol.~4, no.~1, pp. 71--81, 2016.

\bibitem{schwarting2021stochastic}
W.~Schwarting, A.~Pierson, S.~Karaman, and D.~Rus, ``Stochastic dynamic games in belief space,'' \emph{IEEE Transactions on Robotics}, vol.~37, no.~6, pp. 2157--2172, 2021.

\bibitem{pachter2016linear}
M.~Pachter, ``On linear-quadratic gaussian dynamic games,'' in \emph{International Symposium on Dynamic Games and Applications}.\hskip 1em plus 0.5em minus 0.4em\relax Springer, 2016, pp. 301--322.

\bibitem{pachter2017lqg}
------, ``Lqg dynamic games with a control-sharing information pattern,'' \emph{Dynamic Games and Applications}, vol.~7, no.~2, pp. 289--322, 2017.

\bibitem{vasal2016signaling}
D.~Vasal and A.~Anastasopoulos, ``Signaling equilibria for dynamic lqg games with asymmetric information,'' in \emph{2016 IEEE 55th Conference on Decision and Control (CDC)}.\hskip 1em plus 0.5em minus 0.4em\relax IEEE, 2016, pp. 6901--6908.

\bibitem{heydaribeni2019linear}
N.~Heydaribeni and A.~Anastasopoulos, ``Linear equilibria for dynamic lqg games with asymmetric information and dependent types,'' in \emph{2019 IEEE 58th Conference on Decision and Control (CDC)}.\hskip 1em plus 0.5em minus 0.4em\relax IEEE, 2019, pp. 5971--5976.

\bibitem{hambly2023linear}
B.~Hambly, R.~Xu, and H.~Yang, ``Linear-quadratic gaussian games with asymmetric information: Belief corrections using the opponents actions,'' \emph{arXiv preprint arXiv:2307.15842}, 2023.

\bibitem{guan2026forward}
Y.~Guan, I.~Shames, and T.~Summers, ``Forward-backward dynamic programming for lqg dynamic games with partial and asymmetric information,'' \emph{arXiv preprint arXiv:2603.18304}, 2026.

\bibitem{guan2025best}
------, ``Best response convergence for zero-sum stochastic dynamic games with partial and asymmetric information,'' \emph{arXiv preprint arXiv:2501.06181}, 2025.

\bibitem{tzoumas2020lqg}
V.~Tzoumas, L.~Carlone, G.~J. Pappas, and A.~Jadbabaie, ``Lqg control and sensing co-design,'' \emph{IEEE Transactions on Automatic Control}, vol.~66, no.~4, pp. 1468--1483, 2020.

\bibitem{zhong2024sparse}
Y.~Zhong, N.~Yang, L.~Huang, G.~Shi, and L.~Shi, ``Sparse sensor selection for distributed systems: An l1-relaxation approach,'' \emph{Automatica}, vol. 165, p. 111670, 2024.

\bibitem{fotiadis2024input}
F.~Fotiadis and K.~G. Vamvoudakis, ``Input-output data-driven sensor selection,'' in \emph{2024 IEEE 63rd Conference on Decision and Control (CDC)}.\hskip 1em plus 0.5em minus 0.4em\relax IEEE, 2024, pp. 4506--4511.

\bibitem{fotiadis2026input}
------, ``Input--output data-driven sensor selection for cyber-physical systems,'' \emph{Automatica}, vol. 186, p. 112829, 2026.

\bibitem{kassir2015communication}
A.~Kassir, R.~Fitch, and S.~Sukkarieh, ``Communication-aware information gathering with dynamic information flow,'' \emph{The International Journal of Robotics Research}, vol.~34, no.~2, pp. 173--200, 2015.

\bibitem{lauri2017periodic}
M.~Lauri, E.~Hein{\"a}nen, and S.~Frintrop, ``Multi-robot active information gathering with periodic communication,'' in \emph{2017 IEEE International Conference on Robotics and Automation (ICRA)}.\hskip 1em plus 0.5em minus 0.4em\relax IEEE, 2017, pp. 851--856.

\bibitem{best2019decmcts}
G.~Best, O.~M. Cliff, T.~Patten, R.~R. Mettu, and R.~Fitch, ``{Dec-MCTS}: Decentralized planning for multi-robot active perception,'' \emph{The International Journal of Robotics Research}, vol.~38, no. 2--3, pp. 316--337, 2019.

\bibitem{bacsar1998dynamic}
T.~Ba{\c{s}}ar and G.~J. Olsder, \emph{Dynamic noncooperative game theory}.\hskip 1em plus 0.5em minus 0.4em\relax SIAM, 1998.

\bibitem{tassa2014control}
Y.~Tassa, N.~Mansard, and E.~Todorov, ``Control-limited differential dynamic programming,'' in \emph{2014 IEEE International Conference on Robotics and Automation (ICRA)}.\hskip 1em plus 0.5em minus 0.4em\relax IEEE, 2014, pp. 1168--1175.

\bibitem{li2004iterative}
W.~Li and E.~Todorov, ``Iterative linear quadratic regulator design for nonlinear biological movement systems,'' in \emph{First International Conference on Informatics in Control, Automation and Robotics}, vol.~2.\hskip 1em plus 0.5em minus 0.4em\relax SciTePress, 2004, pp. 222--229.

\bibitem{fridovich2020efficient}
D.~Fridovich-Keil, E.~Ratner, L.~Peters, A.~D. Dragan, and C.~J. Tomlin, ``Efficient iterative linear-quadratic approximations for nonlinear multi-player general-sum differential games,'' in \emph{2020 IEEE international conference on robotics and automation (ICRA)}.\hskip 1em plus 0.5em minus 0.4em\relax IEEE, 2020, pp. 1475--1481.

\bibitem{laine2023computation}
F.~Laine, D.~Fridovich-Keil, C.-Y. Chiu, and C.~Tomlin, ``The computation of approximate generalized feedback nash equilibria,'' \emph{SIAM Journal on Optimization}, vol.~33, no.~1, pp. 294--318, 2023.

\bibitem{SmoothGameTheory}
\BIBentryALTinterwordspacing
D.~Fridovich-Keil, \emph{Smooth Game Theory}, 2024. [Online]. Available: \url{https://clearoboticslab.github.io/documents/smooth_game_theory.pdf}
\BIBentrySTDinterwordspacing

\bibitem{welch1995introduction}
G.~Welch and G.~Bishop, ``An introduction to the kalman filter,'' 1995.

\bibitem{beck2009fast}
A.~Beck and M.~Teboulle, ``A fast iterative shrinkage-thresholding algorithm for linear inverse problems,'' \emph{SIAM journal on imaging sciences}, vol.~2, no.~1, pp. 183--202, 2009.

\bibitem{busetto2026nonlinear}
R.~Busetto, E.~Cereda, M.~Forgione, G.~Maroni, D.~Piga, and D.~Palossi, ``Nonlinear system identification for a nano-drone benchmark,'' \emph{Control Engineering Practice}, vol. 172, p. 106871, 2026.

\bibitem{todorov2012mujoco}
E.~Todorov, T.~Erez, and Y.~Tassa, ``Mujoco: A physics engine for model-based control,'' in \emph{2012 IEEE/RSJ international conference on intelligent robots and systems}.\hskip 1em plus 0.5em minus 0.4em\relax IEEE, 2012, pp. 5026--5033.

\end{thebibliography}

\vfill

\end{document}